%% file: iclr2027_conference.tex
\documentclass{article} 
\usepackage{iclr2027_conference,times}

\input{math_commands.tex}

\usepackage{hyperref}
\usepackage{url}
\usepackage{booktabs}
\usepackage{algorithm}
\usepackage{algpseudocode}
\usepackage{xcolor}
\usepackage{booktabs}
\usepackage{tabularx}
\usepackage{graphicx}
\definecolor{fitblue}{RGB}{45,105,170}
\definecolor{calorange}{RGB}{205,120,35}
\definecolor{allocgreen}{RGB}{50,135,85}
\definecolor{decodepurple}{RGB}{120,75,155}
\definecolor{commentgray}{RGB}{120,120,120}

\usepackage[table]{xcolor}
\definecolor{standardblue}{HTML}{EAF2FF}
\definecolor{largeorange}{HTML}{FFF0E6}

\makeatletter
{%
    \kern 2pt
    \hrule
    \end{center}
}
\makeatother

\algrenewcommand\algorithmiccomment[1]{%
    \hfill\textcolor{commentgray}{$\triangleright$~#1}
}

\title{CaliSplat: Sparse Gaussian Fitting and Calibrated Rate Allocation for Image Compression}

\author{
Zhenbang Zhang, Zhengquan Luo, Kai Zhao \& Zhiqiang Xu \\
Mohamed bin Zayed University of Artificial Intelligence \\
\texttt{Zhenbang.Zhang@mbzuai.ac.ae, Zhengquan.Luo@mbzuai.ac.ae} \\
\texttt{Kai.Zhao@mbzuai.ac.ae, Zhiqiang.Xu@mbzuai.ac.ae}
}

\iclrfinalcopy 

\begin{document}

\maketitle
\lhead{Preprint} 

\begin{abstract}
2D Gaussian representations provide an explicit and lightweight alternative for image representation, enabling fast deterministic decoding without a heavy neural synthesis network.
However, existing per-image Gaussian codecs still require costly image-specific optimization to determine how limited Gaussian capacity should be distributed across heterogeneous regions. We introduce CaliSplat, an optimization-efficient Gaussian image codec that selectively explores candidate representations and globally allocates a constrained Gaussian budget.
We exploit the closed-form reconstruction error of the single-Gaussian case as an inexpensive measure of local difficulty, fitting higher-capacity candidates only in regions where additional Gaussians are likely to be useful.
To reduce selection errors caused by mismatched convergence rates across candidate sizes, a small set of longer-horizon fits is used to calibrate short-horizon distortion estimates before global budget-aware selection.
The selected representation is then compressed into a bitstream.
Across standard and large image datasets, CaliSplat achieves competitive or superior rate--distortion performance while reducing per-image fitting time by $3.0\times$--$15.7\times$ compared with representative Gaussian codecs.
It further maintains high-throughput deterministic decoding at $175.9$--$478.2$~MP/s.
These results demonstrate that content-adaptive Gaussian compression can substantially reduce per-image optimization cost while preserving strong reconstruction quality and efficient explicit decoding.
\end{abstract}

\section{Introduction}

\begin{figure}[t]
    \centering
    \includegraphics[width=0.98\linewidth]{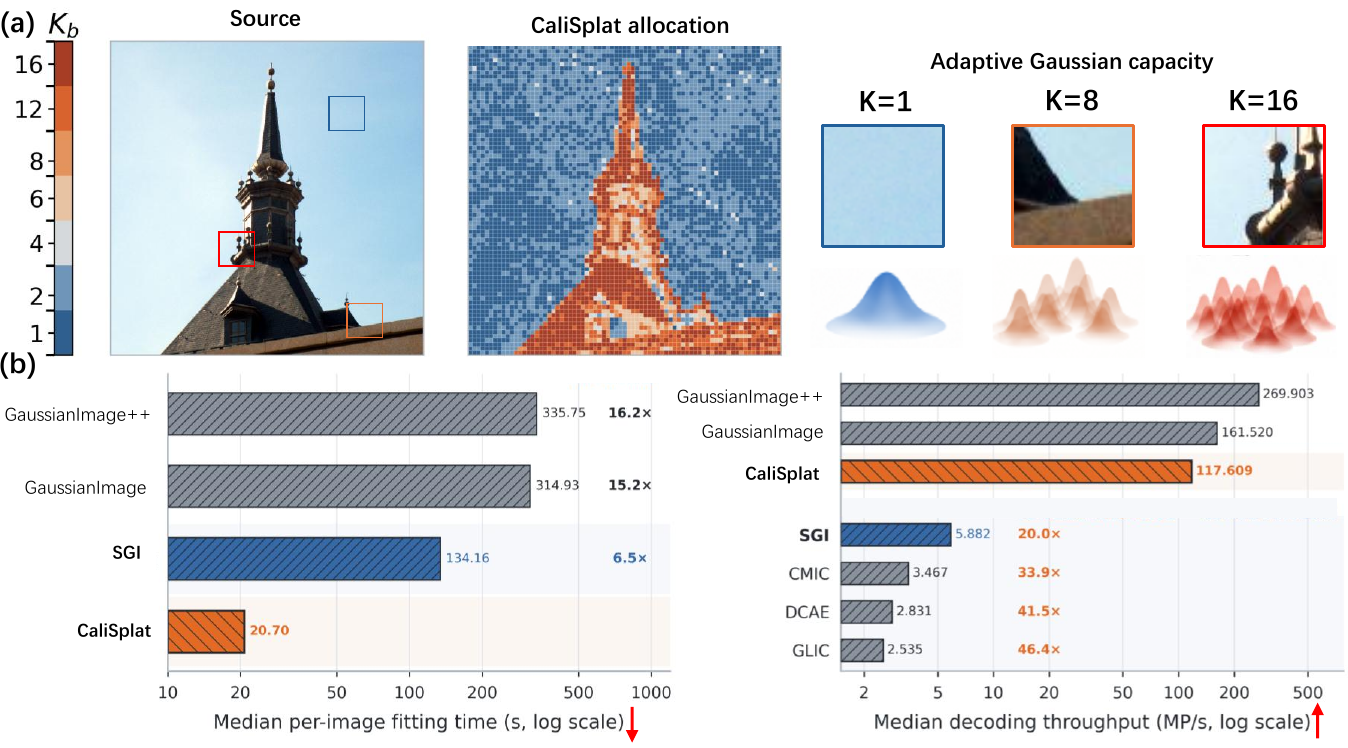}
    \caption{
    \textbf{(a)} Content-adaptive Gaussian allocation assigns larger Gaussian capacity to more complex regions.
    \textbf{(b)} CaliSplat achieves fast fitting while maintaining high decoding throughput.
    }
    \label{fig:head}
\end{figure}

Image compression aims to represent visual signals compactly while preserving reconstruction quality and enabling efficient decoding.
Modern learned codecs achieve strong rate--distortion performance through expressive transforms and entropy models
\citep{balle2016end,balle2018variational,minnen2018joint,cheng2020learned},
while per-image approaches optimize compact representations for individual inputs and can employ substantially lighter decoders
\citep{dupont2021coin,dupont2022coinpp,ladune2023coolchic,kim2024c3}.
More recently, 2D Gaussian Splatting has emerged as an attractive explicit alternative for per-image representation and compression
\citep{zhang2024gaussianimage,zhu2025lig,zeng2025instantgaussianimage,pan2026sgi}.
By reconstructing images from explicit Gaussian primitives through highly parallel rasterization, these methods offer lightweight and high-throughput decoding.

Despite these advantages, efficient Gaussian compression remains challenging because natural images require highly non-uniform representation capacity.
GaussianImage establishes 2D Gaussian primitives for image compression
\citep{zhang2024gaussianimage},
while LIG introduces hierarchical Gaussian representations
\citep{zhu2025lig},
Instant GaussianImage accelerates fitting through learned initialization and adapts the overall Gaussian count
\citep{zeng2025instantgaussianimage},
and SGI improves structure and optimization through seed-based generation and coarse-to-fine fitting
\citep{pan2026sgi}.
However, these methods do not explicitly address how a limited Gaussian budget should be distributed across spatially heterogeneous regions.
Smooth areas may require only a few Gaussians, whereas edges and textures benefit from substantially larger local capacity.
Determining such fine-grained allocation naively requires fitting many candidate representations that are ultimately discarded.
This creates a central trade-off between content-adaptive capacity allocation and per-image optimization efficiency.

We introduce CaliSplat, a per-image Gaussian codec that addresses this trade-off through sparse candidate fitting and global budget-aware allocation.
Our key observation is that the single-Gaussian case admits a closed-form solution: under squared error, its distortion $D_b(1)$ can be computed analytically and serves as an inexpensive measure of local representation difficulty.
We therefore fit higher-capacity candidates only in regions likely to benefit from additional Gaussians, substantially reducing unnecessary optimization.
Since short-horizon distortions may mis-rank candidates with different convergence rates, a small set of longer-horizon fits is used to calibrate their scores before global selection.
The resulting scores guide image-level allocation of a constrained Gaussian budget, concentrating capacity in structurally complex regions as illustrated in Fig.~\ref{fig:head}(a).
Finally, the selected Gaussians are compressed into a bitstream using deterministic geometric templates and structured template-relative parameter coding.

Experiments demonstrate both the component effectiveness and end-to-end efficiency of CaliSplat.
The analytic difficulty gate reduces the $K$-weighted candidate-fitting workload to $65\%$ of exhaustive search with essentially unchanged reconstruction quality, while budget-aware allocation improves PSNR by $1.20$~dB over uniform allocation and by $2.92$~dB over a matched spatially shuffled baseline. On the main datasets, CaliSplat achieves competitive rate--distortion performance on standard images and superior PSNR on large-image datasets, while reducing per-image fitting time by $3.0\times$--$15.7\times$ compared with representative Gaussian codecs.
As shown in Fig.~\ref{fig:head}(b), it additionally retains high-throughput deterministic decoding at $175.9$--$478.2$~MP/s.

Our contributions are summarized as follows:
\begin{itemize}
    \item We introduce an analytic difficulty-guided sparse fitting strategy that leverages the closed-form distortion of a single Gaussian to identify regions that require higher representation capacity, substantially reducing redundant per-image optimization.

    \item We propose a calibrated, budget-aware Gaussian allocation framework that calibrates inexpensive short-horizon candidate evaluations with a small set of longer-horizon fits, enabling globally optimized capacity allocation under a constrained image-level Gaussian budget.

    \item Extensive experiments demonstrate that CaliSplat achieves competitive or superior rate--distortion performance while substantially improving per-image optimization efficiency and maintaining high decoding throughput, with particularly strong performance in high-rate and high-resolution regimes.
\end{itemize}

\section{Related Work}
\label{app:background}

\subsection{Image Compression}
\label{sec:rw_image_compression}

Learned image compression typically maps images to quantized latents and
reconstructs them with a learned synthesis transform.
Early work established end-to-end rate--distortion optimization,
hyperpriors, autoregressive context modeling, and expressive likelihood
models
\citep{minnen2018joint,cheng2020learned}.
Later methods improve coding efficiency and context modeling through
space--channel grouping and hybrid Transformer--CNN architectures
\citep{he2022elic,liu2023tcm}.
Recent systems further explore linear attention, dictionary-based entropy
models, content-adaptive state-space models, and hierarchical context
modeling
\citep{feng2025lalic,lu2025dictionary,qin2025cassic,li2025hpcm},
while practical and adaptive codec design remains an active direction
\citep{tatwawadi2026practical,chen2026glic,ma2026learned}.

Per-image codecs instead optimize a representation for each test image.
COIN and subsequent INR-based methods encode image-specific
representations
\citep{dupont2021coin,strumpler2022implicit,
ladune2023coolchic}.
C3 further demonstrates strong rate--distortion performance with a
low-complexity overfitted decoder
\citep{kim2024c3}, while recent work studies perceptual objectives and
online image-specific optimization
\citep{balle2025good,kuang2025cgoo}.
GaussianImage introduced 2D Gaussian splatting for image representation
and compression
\citep{zhang2024gaussianimage}.
LIG explore content-adaptive or multilevel Gaussian
representations
\citep{zhu2025lig}, while Instant GaussianImage reduces
fitting cost and adapts the primitive count to image content
\citep{zeng2025instantgaussianimage}.

\subsection{3D Gaussian Splatting Compression}
\label{sec:rw_3dgs_compression}

3D Gaussian Splatting represents radiance fields using explicit anisotropic primitives and differentiable rasterization
\citep{kerbl2023gaussians}.
Because high-quality scenes often contain a large number of Gaussians with geometry, opacity, and appearance attributes, compression methods have explored pruning, masking, vector quantization, and compact parameter representations
\citep{niedermayr2024compressed,fan2024lightgaussian}.
Subsequent work further exploits spatial dependencies through structured representations and contextual entropy models
\citep{lu2024scaffold,wang2024contextgs},
while MesonGS studies efficient post-training coding
\citep{xie2024mesongs}.
Recent methods extend these directions with feed-forward compression
\citep{chen2025fcgs},
feature-plane coding with standard video codecs
\citep{lee2025featureplane},
and spatial-distribution optimization
\citep{yang2026gs2}.
These studies highlight that Gaussian compression benefits not only from compact attribute coding, but also from structured representations and adaptive allocation of limited Gaussian capacity.

\section{Proposed Method: CaliSplat}
\label{sec:method}

We introduce \textbf{CaliSplat}, a per-image compression framework based on explicit anisotropic Gaussians.
Given an RGB image $I$, the encoder produces a bitstream $S$, from which the decoder recovers an explicit Gaussian representation $\widehat{\Theta}$ and deterministically reconstructs the image:
\begin{equation}
    \widehat{\Theta}
    =
    \operatorname{Dec}_{\Theta}(S),
    \qquad
    \widehat I
    =
    \mathcal R(\widehat{\Theta}).
    \label{eq:overview}
\end{equation}

\begin{figure}[t]
    \centering
    \includegraphics[width=\linewidth]{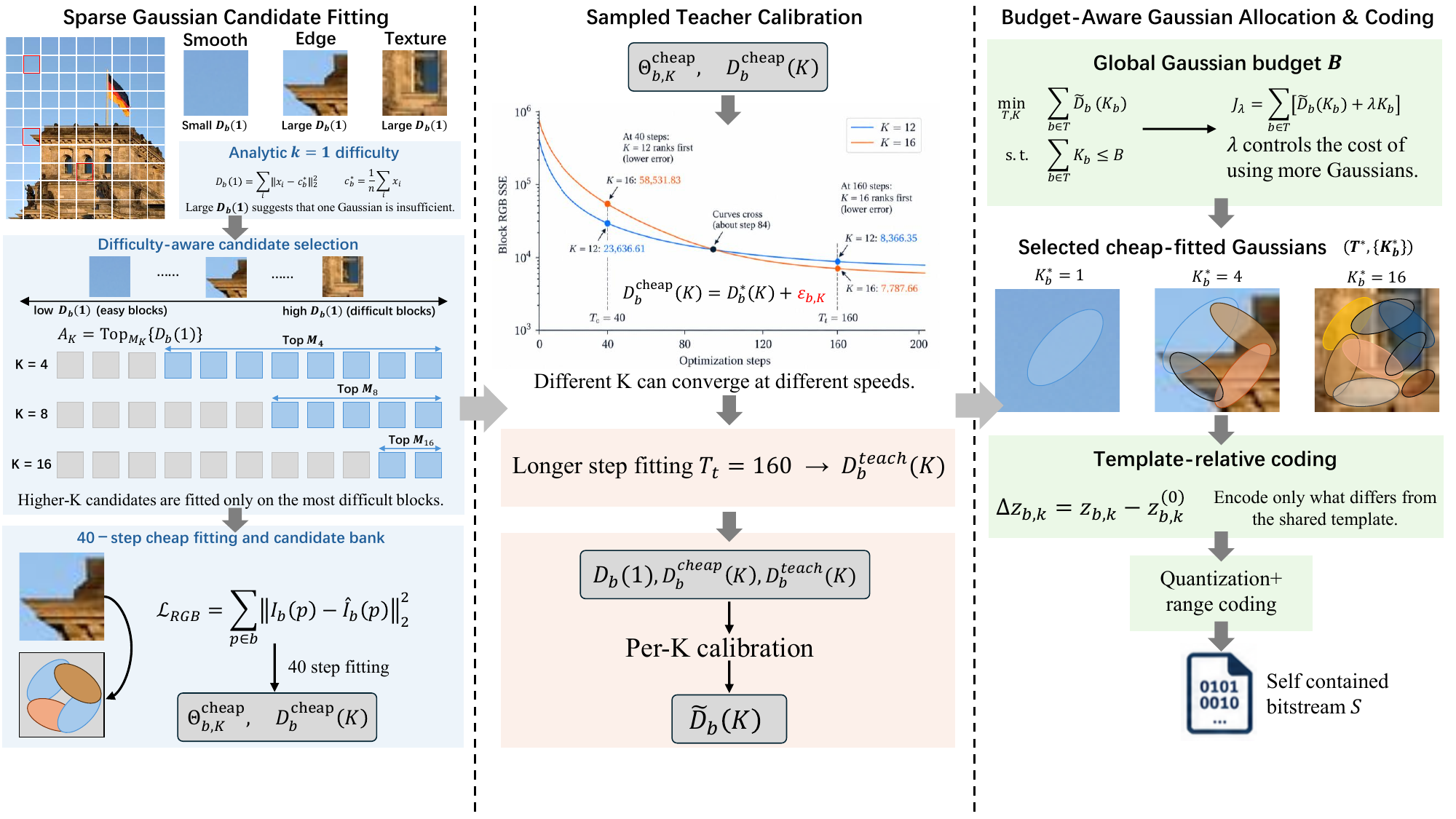}
    \vspace{-20pt}
\caption{The overview of the CaliSplat encoding pipeline.
Sparse candidate fitting reduces unnecessary optimization, sampled teacher calibration corrects short-horizon scores, and budget-aware allocation selects and codes the final Gaussian representation.}
    \label{fig:encoder_overview}
\end{figure}

The main challenge is to determine where Gaussian should be spent without fully optimizing every possible representation.
As shown in Fig.~\ref{fig:encoder_overview}, we first construct sparse candidates only where additional Gaussian is likely to be useful (Sec.~\ref{sec:sparse}).
Since short optimization can favor candidate sizes that simply converge faster, we use a small number of longer-horizon reference fits to calibrate their scores (Sec.~\ref{sec:calibration}).
These calibrated scores guide the global Gaussian-budget allocation, after which the selected parameters are compactly quantized and entropy-coded into the bitstream $S$ (Sec.~\ref{sec:coding}).
The decoding only recovers the explicit Gaussian parameters from $S$ and rasterizes them into the reconstructed image (Sec.~\ref{sec:decoder}).
Alg.~\ref{alg:encoding} and~\ref{alg:decoding} summarize the complete encoding and decoding procedures, respectively.

\subsection{Sparse Gaussian Candidate Fitting}
\label{sec:sparse}

\paragraph{Local Gaussian representation.}
We partition the image into many blocks and represent each block with a variable number $K$ of local anisotropic Gaussians. The smooth regions can often be represented with few Gaussians, whereas edges and textures may require more.
A Gaussian in block $b$ is parameterized as
\begin{equation}
    \theta_{b,k}
    =
    \left(
        \boldsymbol\mu_{b,k},
        \boldsymbol s_{b,k},
        \phi_{b,k},
        a_{b,k},
        \boldsymbol c_{b,k}
    \right),
\end{equation}
where $\boldsymbol\mu$, $\boldsymbol s=(s_1,s_2)$, $\phi$, $a$, and $\boldsymbol c$ denote center, principal-axis scales, orientation, log-weight, and RGB color, respectively.
We denote a $K$-Gaussian candidate by
$\Theta_{b,K}=\{\theta_{b,k}\}_{k=1}^{K}$.

For a local coordinate $\mathbf p$, the Gaussian logit is
\begin{equation}
    \ell_{b,k}(\mathbf p)
    =
    -\frac12
    \left[
        \left(
            \frac{d_{1,b,k}(\mathbf p)}
            {s_{1,b,k}}
        \right)^2
        +
        \left(
            \frac{d_{2,b,k}(\mathbf p)}
            {s_{2,b,k}}
        \right)^2
    \right]
    +a_{b,k},
    \label{eq:logit}
\end{equation}
where $d_1$ and $d_2$ are coordinates along the rotated principal axes.
The normalized Gaussian weights and block reconstruction are
\begin{equation}
    \pi_{b,k}(\mathbf p)
    =
    \frac{\exp(\ell_{b,k}(\mathbf p))}
    {\sum_{j=1}^{K}\exp(\ell_{b,j}(\mathbf p))},
    \qquad
    \widehat I_b(\mathbf p)
    =
    \sum_{k=1}^{K}
    \pi_{b,k}(\mathbf p)\boldsymbol c_{b,k}.
    \label{eq:rasterization}
\end{equation}
Thus, each pixel is reconstructed as a local mixture of Gaussian colors without depth ordering or alpha compositing.
Although a larger $K$ provides more fitting capacity, optimizing every possible $K$ for every block would make per-image encoding expensive.
We therefore first ask a cheaper question: which blocks are likely to need more than the simplest representation?

\begin{figure}[t]
    \centering
    \includegraphics[width=\linewidth]{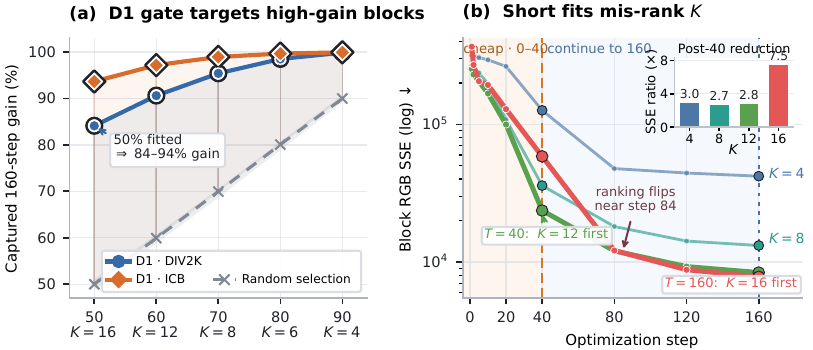}
    \vspace{-20pt}
    \caption{
    (a) The analytic D1 gate preserves most long-horizon fitting gain while evaluating only a subset of candidates. DIV2K and ICB denote the datasets used for these analyses.
    (b) A real order-reversal example shows that short fitting can mis-rank different $K$ due to convergence mismatch, motivating sampled teacher calibration.
    }
    \label{fig:gate-and-convergence-mismatch}
\end{figure}

\paragraph{Analytic difficulty estimation.}
We answer this question using the $K=1$ case, which requires no iterative optimization.
With a single Gaussian, the softmax weight is identically one, so the entire block reduces to one constant RGB color.
Under squared error, the optimal color is simply the block mean:
\begin{equation}
    \boldsymbol c_b^\star
    =
    \frac1n
    \sum_{i=1}^{n}
    \mathbf x_i,
    \qquad
    D_b(1)
    =
    \sum_{i=1}^{n}
    \left\|
        \mathbf x_i-\boldsymbol c_b^\star
    \right\|_2^2.
    \label{eq:k1}
\end{equation}
This gives an intuitive difficulty signal at essentially zero fitting cost.
If $D_b(1)$ is small, a single color already explains the block well; if it is large, the block may benefit from additional Gaussians.
We therefore use $D_b(1)$ to decide where more expensive candidates are worth trying.

\paragraph{Difficulty-aware sparse fitting.}
For each candidate size $K$, we rank all image blocks by the analytic
difficulty $D_b(1)$ and retain the top $M_K$ blocks:
\begin{equation}
    \mathcal A_K
    =
    \operatorname{Top}_{M_K}
    \left\{
        D_b(1)
    \right\}_{b},
    \label{eq:gate}
\end{equation}
where $M_K$ is determined by a predefined retention fraction.
This gate $M_K$  is a computational filter: it decides where optimization effort
is spent, while the final Gaussian count is chosen later by the global
allocator.
As shown in Fig.~\ref{fig:gate-and-convergence-mismatch}(a),
this simple ranking preserves most of the long-horizon fitting gain while
evaluating only a subset of candidates.

For each admitted pair $(b,K)$, $\theta_{b,k}$ is initialized from a deterministic template defined by the block size $r_b$ and Gaussian count $K$, and colors are initialized in closed form as
\begin{equation}
    C^{(0)}
    =
    W^\dagger X,
    \label{eq:ls_init}
\end{equation}
where $W$ is the template responsibility matrix and $X$ is the flattened RGB block.
Starting from this initialization, we perform only $T_c=40$ Adam steps.
At each step, the current Gaussians are rasterized using Eq.~\ref{eq:rasterization}, their RGB reconstruction error is computed against the source block, and gradients are backpropagated through the differentiable rasterizer.
The resulting short-horizon candidate is
\begin{equation}
    \Theta_{b,K}^{\mathrm{cheap}},
    \qquad
    D_b^{\mathrm{cheap}}(K).
    \label{eq:cheap}
\end{equation}
Candidates outside $\mathcal A_K$ are simply skipped, so they incur no fitting cost.
For the admitted candidates, the short optimization produces a sparse bank of cheap-fitted Gaussian representations
$\Theta_{b,K}^{\mathrm{cheap}}$
and their corresponding distortions
$D_b^{\mathrm{cheap}}(K)$.
Template construction, retention fractions, and optimization details are provided in Appendices~\ref{app:param} and~\ref{app:sparse}.

\subsection{Sampled Teacher Calibration}
\label{sec:calibration}

The sparse fitting stage in Sec.~\ref{sec:sparse} efficiently produces
multiple candidates for each block.
However, different candidate sizes can converge at different speeds, so
their errors after only $T_c=40$ steps may not preserve their longer-horizon
ranking.
Fig.~\ref{fig:gate-and-convergence-mismatch}(b) shows a real order-reversal
example: $K=12$ has lower error at 40 steps, whereas $K=16$ becomes better
after longer optimization.
We therefore use longer-horizon (teachers) fits as reference signals
for correcting the short-fit scores.

\paragraph{Short-horizon optimization bias.}
We write
\begin{equation}
    D_b^{\mathrm{cheap}}(K)
    =
    D_b^\star(K)
    +
    \varepsilon_{b,K},
    \qquad
    \varepsilon_{b,K}\ge0,
    \label{eq:bias}
\end{equation}
where $D_b^\star(K)$ denotes the distortion after sufficiently long optimization and
$\varepsilon_{b,K}$ is the remaining error caused by stopping early.

\paragraph{Sampled teacher fitting.}
We run teacher optimization only on a small representative subset.
For each $K>1$, we sample at most $128$ admitted candidates across the range of $D_b^{\mathrm{cheap}}(K)$.
Each sampled candidate is reinitialized from the same deterministic template and optimized for $T_t=160$ steps, producing the teacher error $D_b^{\mathrm{teach}}(K)$.
The sampling rule is detailed in Appendix~\ref{app:calibration}.

\paragraph{Per-$K$ distortion calibration.}
The teacher errors provide longer-horizon references for correcting the 40-step candidate scores.
For each candidate, we construct
\begin{equation}
    \mathbf{x}_{b,K}
    =
    \left[
    1,\,
    \log\!\left(1+D_b^{\mathrm{cheap}}(K)\right),\,
    \log\!\left(1+D_b(1)\right)
    \right]^{\top},
    \label{eq:cal_feat}
\end{equation}
where $D_b^{\mathrm{cheap}}(K)$ describes the observed short fit and
$D_b(1)$ provides a simple measure of the block's intrinsic difficulty.
Since different values of $K$ can have different convergence behavior, we fit a separate least-squares mapping for each $K$ to predict the corresponding teacher distortion.
This gives the calibrated score $\widetilde D_b(K)$, with
$\widetilde D_b(1)=D_b(1)$.

\subsection{Budget-Aware Gaussian Coding}
\label{sec:coding}

After calibration in Sec.~\ref{sec:calibration}, each block has several candidate representations with different Gaussian counts $K$.
The remaining question is how to choose among these candidates when the entire image is allowed to use only a limited number of Gaussians.

\paragraph{Gaussian-budget allocation.}
A larger $K$ usually gives a block more fitting capacity, but spending more Gaussians on one block leaves fewer for the others.
We therefore allocate a global budget $B$ by jointly choosing the image partition $\mathcal T$ and the Gaussian count $K_b$ for each selected block:
\begin{equation}
    \min_{\mathcal T,\mathbf K}
    \quad
    \sum_{b\in\mathcal T}
    \widetilde D_b(K_b)
    \qquad
    \text{s.t.}\quad
    \sum_{b\in\mathcal T}
    K_b
    \le B.
    \label{eq:budget}
\end{equation}
Intuitively, the allocator spends more Gaussians on blocks where additional capacity reduces reconstruction error the most, and fewer on blocks that are already represented well with small $K$.

Directly enforcing the global budget while choosing both the block structure and $K$ is difficult.
We assign a cost $\lambda$ to each Gaussian and minimize
\begin{equation}
    \mathcal J_\lambda
    =
    \sum_{b\in\mathcal T}
    \left[
        \widetilde D_b(K_b)
        +
        \lambda K_b
    \right].
    \label{eq:lagrangian}
\end{equation}
A smaller $\lambda$ allows more Gaussians, while a larger $\lambda$ favors more compact representations.
We adjust $\lambda$ until the total number of selected Gaussians is close to the target budget $B$, and retain the corresponding cheap-fitted candidates
$\Theta_{b,K}^{\mathrm{cheap}}$.
The detailed allocation procedure is given in Appendix~\ref{app:allocation}.

\paragraph{Compression-aware parameter coding.}
Once the allocation is fixed, only the selected Gaussian parameters need to be transmitted.
We encode the fitted geometry relative to the decoder-known template:
\begin{equation}
    \Delta\mathbf z_{b,k}
    =
    \mathbf z_{b,k}
    -
    \mathbf z^{(0)}_{b,k},
    \label{eq:geom_residual}
\end{equation}
where $\mathbf z^{(0)}_{b,k}$ and $\mathbf z_{b,k}$ denote the template and fitted geometry, respectively.
The resulting geometry residuals, relative weights, colors, topology, and Gaussian counts are quantized and losslessly entropy-coded with a standard range coder
\citep{witten1987arithmetic},
forming the self-contained bitstream $S$.
Detailed coding settings are provided in Appendix~\ref{app:implementation}.

\subsection{Deterministic Gaussian Decoding}
\label{sec:decoder}

Since the bitstream stores an explicit Gaussian representation, decoding only requires entropy decoding, inverse quantization, and rasterization.
The decoded block size $r_b$ and Gaussian count $K_b$ determine the shared geometric template, from which the fitted geometry is reconstructed as
\begin{equation}
    \widehat{\mathbf z}_{b,k}
    =
    \mathbf z^{(0)}_{b,k}(r_b,K_b)
    +
    Q_z^{-1}\!\left(\mathbf q^{z}_{b,k}\right).
    \label{eq:decode_geometry}
\end{equation}
The remaining attributes are recovered analogously, yielding
$\widehat{\Theta}$, which is rasterized using Eq.~\ref{eq:rasterization} to obtain the final reconstruction $\widehat I$.
Details are provided in Appendix~\ref{app:decoder}.

\section{Experiments}
\label{sec:experiments}

\subsection{Experimental Setup}
\label{sec:exp_setup}

\paragraph{Datasets and Baselines.}
We evaluate CaliSplat on two groups of datasets covering both standard-resolution and large images.
The standard-resolution set contains 75 images, with 25 images each from
Kodak~\citep{toderici2017full},
Tecnick~\citep{asuni2013testimages},
and CLIC2020~\citep{cheng2020learned}.
The large-image set contains 9 images, with 3 images each from
FGF2~\citep{wang2024cross},
ICB~\citep{ororbia2019learned},
and STimage~\citep{chen2024stimage}.
We compare with representative Gaussian-based methods, including GaussianImage \citep{zhang2024gaussianimage}, GaussianImage++ \citep{li2026gaussianimagepp}, Image-GS \citep{zhang2024imagegs}, SGI \citep{pan2026sgi}, and LIG \citep{zhu2025lig}.

\paragraph{Implementation details.}
We use $16\times16$ root blocks with candidate Gaussian counts
$K\in\{1,2,4,6,8,12,16\}$.
Each admitted candidate is optimized for 40 Adam steps with a learning rate of
$0.035$, while the sampled teacher fits use 160 steps and at most 128 samples
for each $K$.
Additional parameterization and coding settings are provided in
Appendix~\ref{app:implementation}.

\subsection{Results and Analysis}
\label{sec:exp_results}

\begin{table}[]
\centering
\resizebox{\textwidth}{!}{%
\begin{tabular}{@{}lrrrrrrrr@{}}
\toprule

& \multicolumn{4}{c}{\cellcolor{standardblue}\textbf{Low-rate}}
& \multicolumn{4}{c}{\cellcolor{largeorange}\textbf{High-rate}} \\

\cmidrule(lr){2-5}
\cmidrule(lr){6-9}

\multicolumn{1}{c}{\textbf{Method}} &
\multicolumn{1}{c}{\cellcolor{standardblue}\textbf{\begin{tabular}[c]{@{}c@{}}PSNR\\(dB) $\uparrow$\end{tabular}}} &
\multicolumn{1}{c}{\cellcolor{standardblue}\textbf{MS-SSIM $\uparrow$}} &
\multicolumn{1}{c}{\cellcolor{standardblue}\textbf{\begin{tabular}[c]{@{}c@{}}Training time\\(s) $\downarrow$\end{tabular}}} &
\multicolumn{1}{c}{\cellcolor{standardblue}\textbf{\begin{tabular}[c]{@{}c@{}}Decode speed\\(MP/s) $\uparrow$\end{tabular}}} &
\multicolumn{1}{c}{\cellcolor{largeorange}\textbf{\begin{tabular}[c]{@{}c@{}}PSNR\\(dB) $\uparrow$\end{tabular}}} &
\multicolumn{1}{c}{\cellcolor{largeorange}\textbf{MS-SSIM $\uparrow$}} &
\multicolumn{1}{c}{\cellcolor{largeorange}\textbf{\begin{tabular}[c]{@{}c@{}}Training time\\(s) $\downarrow$\end{tabular}}} &
\multicolumn{1}{c}{\cellcolor{largeorange}\textbf{\begin{tabular}[c]{@{}c@{}}Decode speed\\(MP/s) $\uparrow$\end{tabular}}} \\

\midrule

\multicolumn{9}{c}{\textbf{Standard images}} \\
\midrule

GaussianImage
& 26.21
& 0.8829
& 332.12
& 213.58
& 30.47
& 0.9596
& 345.42
& 129.41 \\

GaussianImage++
& 26.30
& 0.8853
& 341.99
& \textbf{256.53}
& 30.30
& 0.9578
& 346.98
& \textbf{241.74} \\

Image-GS
& \textbf{27.95}
& \textbf{0.8994}
& \underline{77.50}
& \underline{227.00}
& \underline{32.08}
& \underline{0.9607}
& \underline{71.51}
& \underline{205.51} \\

\textbf{CaliSplat}
& \underline{27.71}
& \underline{0.8866}
& \textbf{22.03}
& 215.82
& \textbf{32.95}
& \textbf{0.9655}
& \textbf{23.84}
& 175.90 \\

\midrule

\multicolumn{9}{c}{\textbf{Large images}} \\
\midrule

LIG
& 23.64
& 0.7104
& \underline{848.69}
& \underline{390.84}
& 29.70
& 0.9163
& \underline{1015.55}
& \textbf{374.36} \\

SGI
& \underline{26.67}
& \textbf{0.8634}
& 958.89
& 8.49
& \underline{30.55}
& \underline{0.9533}
& 1617.21
& 1.93 \\

\textbf{CaliSplat}
& \textbf{27.10}
& \underline{0.8504}
& \textbf{94.86}
& \textbf{478.19}
& \textbf{31.86}
& \textbf{0.9645}
& \textbf{103.25}
& \underline{255.24} \\

\bottomrule
\end{tabular}%
}

\caption{
Quantitative comparison on standard- and large-image datasets.
We compare with GaussianImage \citep{zhang2024gaussianimage},
GaussianImage++ \citep{li2026gaussianimagepp},
Image-GS \citep{zhang2024imagegs},
LIG \citep{zhu2025lig}, and SGI \citep{pan2026sgi}.
The best and second-best results within each setting are shown in bold
and underlined, respectively.
}
\label{tab:quantitative_comparison}
\end{table}

\begin{figure*}[t]
    \centering
    \includegraphics[width=\linewidth]{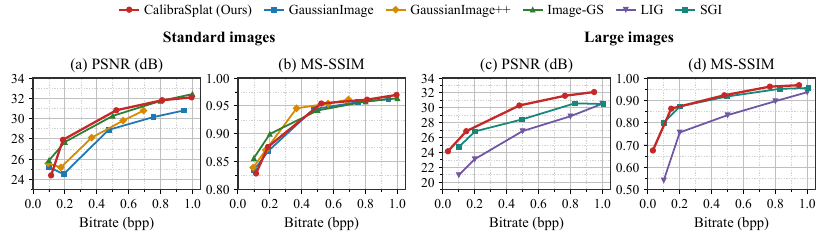}
    \vspace{-20pt}
    \caption{
 Rate-distortion curves of CaliSplat and different baselines on standard images and large images datasets.
    }
    \label{fig:rd_four_panel}
\end{figure*}

\paragraph{Rate--distortion performance.}
Fig.~\ref{fig:rd_four_panel} compares rate--distortion performance on both standard- and large-image datasets.
On standard images, CaliSplat remains competitive across the bitrate range and becomes particularly strong at higher rates.
As shown in Tab.~\ref{tab:quantitative_comparison}, Image-GS achieves slightly higher low-rate PSNR (27.95 vs.\ 27.71~dB), while CaliSplat obtains the best high-rate performance with 32.95~dB PSNR and 0.9655 MS-SSIM.
This trend suggests that the additional representation capacity available at higher rates can be more effectively assigned to regions that benefit from additional Gaussians.

\begin{figure}[t]
    \centering
    \includegraphics[width=\linewidth]{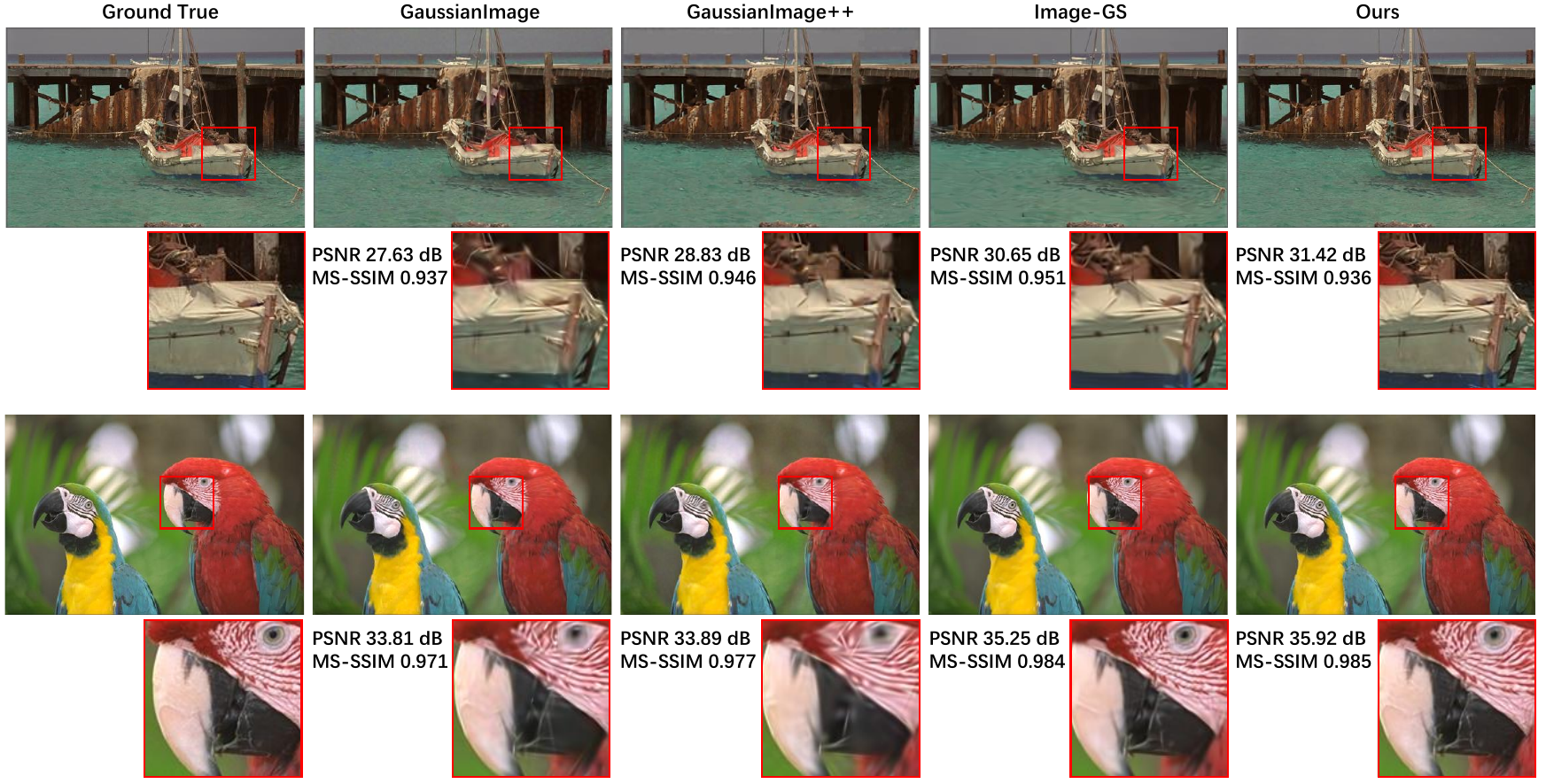}
\caption{
Qualitative reconstruction comparison of CaliSplat with other baselines.
Our method preserves sharper edges and finer local structures with fewer visible artifacts.
}
    \label{fig:vis}
\end{figure}

The advantage becomes more pronounced on large images.
As shown in Fig.~\ref{fig:rd_four_panel} and Tab.~\ref{tab:quantitative_comparison}, CaliSplat achieves the best PSNR at both low and high rates, reaching 27.10 and 31.86~dB, respectively.
At the high-rate point, it outperforms SGI and LIG by 1.31~dB and 2.16~dB, respectively, while also achieving the highest MS-SSIM of 0.9645.
Although SGI performs better in MS-SSIM at the low-rate point, CaliSplat overtakes it as the bitrate increases.
Fig.~\ref{fig:vis} further shows that CaliSplat preserves finer local structures with fewer visible artifacts.
Overall, these results demonstrate that CaliSplat maintains competitive performance on standard images while scaling favorably to larger and more complex content.

\paragraph{Encoding and decoding efficiency.}
CaliSplat also substantially reduces the cost of per-image optimization.
On standard images, Tab.~\ref{tab:quantitative_comparison} reports fitting times of only 22.03--23.84 seconds, corresponding to approximately $3.0\times$--$15.5\times$ faster optimization than the compared Gaussian baselines.
Despite this substantial reduction in fitting cost, CaliSplat maintains high decoding throughput of 175.90--215.82~MP/s, remaining competitive with efficient explicit Gaussian decoders.

The efficiency gain is even more pronounced on large images, where expensive per-image optimization becomes a greater bottleneck.
CaliSplat requires only 94.86--103.25 seconds of fitting, yielding approximately $8.9\times$--$15.7\times$ faster optimization than LIG and SGI, while retaining 255.24--478.19~MP/s decoding throughput.
These results indicate that the efficiency improvement primarily comes from reducing unnecessary encoder-side representation search rather than simplifying the final representation at the expense of reconstruction or decoding efficiency.
Together with the RD results, CaliSplat provides a favorable trade-off between reconstruction quality, per-image fitting cost, and explicit Gaussian decoding efficiency.

\subsection{Ablation Studies}
\label{sec:exp_ablation}

\begin{figure*}[t]
    \centering
    \includegraphics[width=\linewidth]{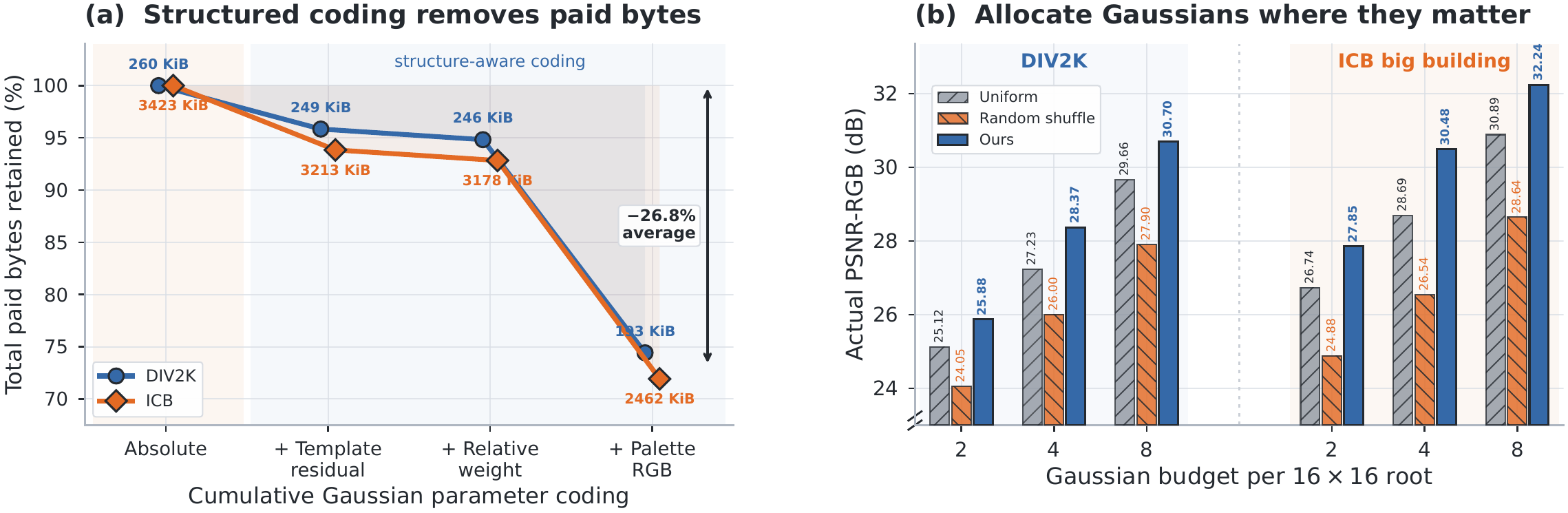}
    \vspace{-20pt}
    \caption{
    Ablation of parameter coding and Gaussian-budget allocation.
    \textbf{(a)} Cumulative coding stages progressively reduce the actual
    transmitted bytes while keeping the selected Gaussians and quantization
    fixed.
    \textbf{(b)} Budget-aware allocation consistently outperforms uniform
    allocation and a histogram-matched random spatial shuffle under the same
    Gaussian budget.
    }
    \label{fig:coding-allocation-ablation}
\end{figure*}

\paragraph{Structured parameter coding.}
Fig.~\ref{fig:coding-allocation-ablation}(a) evaluates the contribution of
our structured parameter coding.
We keep the selected Gaussians, quantization, and entropy coder fixed, and
cumulatively replace absolute coding with template-relative geometry,
relative log-weights, and palette-based RGB coding.
The result reports the resulting bitstream size relative to absolute coding.
The full scheme reduces the stream by $25.57\%$ on DIV2K and $28.07\%$ on
ICB, with the largest additional saving coming from palette-based RGB coding.
These results show that exploiting the structure of the Gaussian
representation substantially reduces the actual transmitted bits without
changing the underlying Gaussian representation.

\paragraph{Budget-aware Gaussian allocation.}
Fig.~\ref{fig:coding-allocation-ablation}(b) evaluates how the available
Gaussian budget should be distributed after candidate fitting.
We compare our method with \emph{Uniform}, which assigns the same $K$ to
all root blocks, and \emph{Random shuffle}, which keeps the same overall
$K$ distribution as our method but randomly moves these counts across
spatial locations.
Across three budgets $B/N\in\{2,4,8\}$, our allocation
outperforms both controls in all six cases.
Compared with Uniform, it improves PSNR by $1.20$ dB on average, showing
that different regions benefit from different Gaussian counts.
Compared with Random shuffle, the average gain increases to $2.92$ dB,
showing that these capacities must also be assigned to the appropriate
spatial regions.

\paragraph{Sparse candidate fitting.}
Tab.~\ref{tab:sparse-fitting-ablation} evaluates which higher-$K$
candidates need to be fitted before Gaussian-budget allocation.
\emph{Exhaustive} fits all supported candidates.
\emph{Random gate} and the \emph{Full Model} use the same reduced fitting
work, but Random gate selects candidates randomly, whereas the Full Model
uses $D_b(1)$ to decide which higher-$K$ candidates are worth fitting.
Compared with exhaustive fitting, the Full Model reduces the
$K$-weighted fitting work from $100\%$ to $65\%$ while slightly improving
PSNR from $30.485$ to $30.509$ dB.
At the same $65\%$ fitting work, random selection gives lower PSNR.
This shows that many candidate fits can be skipped, but the candidates
should be selected using $D_b(1)$ rather than removed at random.

\begin{table}[t]
\centering
\resizebox{0.95\linewidth}{!}{
\begin{tabular}{lcrrrrr}
\toprule
Variant
& Candidates fitted
& \shortstack{Fit work(\%)} $\downarrow$
& \shortstack{Enc.(s)} $\downarrow$
& bpp $\downarrow$
& PSNR $\uparrow$
& SSIM $\uparrow$ \\
\midrule

Exhaustive
& All
& 100.0
& 292.49
& 0.5165
& 30.485
& 0.8901 \\

Random gate
& Random subset
& \textbf{65.0}
& 280.76
& \textbf{0.5162}
& 30.316
& 0.8859 \\

\midrule

\textbf{Full Model}
& $D_b(1)$ subset
& \textbf{65.0}
& \textbf{221.69}
& 0.5164
& \textbf{30.509}
& \textbf{0.8906} \\

\bottomrule
\end{tabular}
}
\caption{
Ablation of sparse candidate fitting
at $B/N=4$.
Fit work denotes the $K$-weighted candidate-fitting work relative to
exhaustive fitting.
Bold values indicate the best result.
}
\label{tab:sparse-fitting-ablation}
\end{table}

\section{Conclusion}

We presented CaliSplat, an efficient per-image Gaussian codec that reduces content-adaptive representation search while preserving lightweight explicit decoding.
By using the closed-form single-Gaussian distortion to guide sparse candidate fitting, followed by calibrated budget-aware allocation, CaliSplat concentrates optimization and Gaussian capacity where they are most useful.
The selected representation is further compressed using deterministic templates and structured template-relative coding.
Experiments on standard and large image datasets show competitive or superior rate--distortion performance with $3.0\times$--$15.7\times$ faster per-image fitting and high-throughput decoding.
These results demonstrate the effectiveness of selective representation search for efficient content-adaptive Gaussian image compression.

\subsection*{AI use statement}

In this work, we used generative AI tools to assist with feedback on
experimental design and the interpretation and presentation of experimental
results.
We additionally used generative AI tools for drafting and editing parts of the
manuscript, improving writing clarity and organization, literature search,
identifying relevant references, and formatting citations.
All AI-assisted technical statements were manually reviewed and verified by the authors.
All reported quantitative results were obtained from author-run experiments
and were not generated or modified by generative AI.
We take responsibility for the final content of this work, including text,
claims, and artifacts produced with the aid of generative AI.

\bibliography{iclr2027_conference}
\bibliographystyle{iclr2027_conference}

\appendix
\section{Appendix}

\begin{algorithm}[t]
\caption{\textbf{Per-Image Encoding with CaliSplat}}
\label{alg:encoding}
\begin{algorithmic}[1]

\Require RGB image $I$; Gaussian budget $B$
\Ensure Self-contained bitstream $S$

\State Partition $I$ into supported spatial blocks

\Statex
\Statex \textcolor{fitblue}{
\textbf{Stage 1: Sparse Gaussian candidate fitting}
\hfill
Sec.~\ref{sec:sparse};
Appendix.~\ref{app:param},~\ref{app:sparse}}

\For{each block $b$}
    \State Compute the analytic $K=1$ solution
    $(\boldsymbol c_b^\star,D_b(1))$
    \Comment{Eq.~\ref{eq:k1}}
\EndFor

\For{each supported $K>1$}
    \State Select admitted blocks $\mathcal A_K$ using $D_b(1)$
    \Comment{Eq.~\ref{eq:gate}}

    \For{each $b\in\mathcal A_K$}
        \State Initialize template geometry and
        $C^{(0)}=W^\dagger X$
        \Comment{Eq.~\ref{eq:ls_init}}

        \For{$t=1,\ldots,T_c$}
            \State Rasterize
            $\widehat I_{b,K}^{(t)}
            =\mathcal R(\Theta_{b,K}^{(t)})$
            \Comment{Eq.~\ref{eq:rasterization}}

            \State Compute
            $\mathcal L_{b,K}^{(t)}
            =
            \sum_{\mathbf p\in b}
            \|I_b(\mathbf p)-\widehat I_{b,K}^{(t)}(\mathbf p)\|_2^2$

            \State Backpropagate
            $\nabla_{\boldsymbol\rho_{b,K}}
            \mathcal L_{b,K}^{(t)}$
            and update $\boldsymbol\rho_{b,K}$ with Adam
        \EndFor

        \State Store
        $\Theta_{b,K}^{\mathrm{cheap}}$
        and $D_b^{\mathrm{cheap}}(K)$
        \Comment{Eq.~\ref{eq:cheap}}
    \EndFor
\EndFor

\Statex
\Statex \textcolor{calorange}{
\textbf{Stage 2: Sampled teacher calibration}
\hfill
Sec.~\ref{sec:calibration};
Appendix.~\ref{app:calibration}}

\For{each $K>1$}
    \State Sample at most $128$ admitted blocks across the
    $D_b^{\mathrm{cheap}}(K)$ range

    \For{each sampled block $b$}
        \State Reinitialize from the same deterministic template
        \State Repeat the same rasterization--loss--backpropagation loop
        for $T_t$ steps
        \State Record $D_b^{\mathrm{teach}}(K)$
    \EndFor

    \State Construct calibration features $\mathbf x_{b,K}$
    \Comment{Eq.~\ref{eq:cal_feat}}

    \State Fit the per-$K$ least-squares mapping and predict
    $\widetilde D_b(K)$ for all admitted candidates
\EndFor

\State Set $\widetilde D_b(1)\gets D_b(1)$
\State Discard teacher parameters and retain
$\Theta_{b,K}^{\mathrm{cheap}}$
\Comment{\textbf{teachers calibrate scores only}}
\Statex
\Statex \textcolor{allocgreen}{
\textbf{Stage 3: Budget-aware Gaussian allocation and coding}
\hfill
Sec.~\ref{sec:coding};
Appendix~\ref{app:allocation}}

\State Search $\lambda$ and solve the hierarchical allocation
using calibrated distortions
\Comment{Eqs.~\ref{eq:budget}--\ref{eq:lagrangian}}

\State Obtain the selected topology $\mathcal T^\star$
and Gaussian counts $\{K_b^\star\}$

\For{each selected leaf $b\in\mathcal T^\star$}
    \State Retrieve the selected
    $\Theta_{b,K_b^\star}^{\mathrm{cheap}}$

    \State Encode template-relative geometry, relative weights,
    and palette-coded colors
    \Comment{Eq.~\ref{eq:geom_residual}}
\EndFor

\State Quantize and range-code the selected topology,
Gaussian counts and parameters into $S$

\State \Return $S$

\end{algorithmic}
\end{algorithm}

\begin{algorithm}[t]
\caption{\textbf{Deterministic Gaussian Decoding}}
\label{alg:decoding}
\begin{algorithmic}[1]

\Require CaliSplat bitstream $S$
\Ensure Reconstructed RGB image $\widehat I$

\Statex
\Statex \textcolor{decodepurple}{\textbf{Stage 1: Recover the explicit Gaussian representation}}
\Statex \textcolor{commentgray}{\hspace{\algorithmicindent}
Sec.~\ref{sec:decoder}; inverse details in Appendix~\ref{app:decoder}}

\State Decode the selected topology, Gaussian counts,
and quantized Gaussian symbols

\For{each selected leaf $b$}
    \State Reconstruct the deterministic template from $(r_b,K_b)$

    \State Recover coding-space geometry from the template
    and transmitted residual
    \Comment{Eq.~\ref{eq:decode_geometry}}

    \State Recover relative weights and Gaussian colors

    \State Assemble the decoded Gaussian set
    $\widehat{\Theta}_b$
\EndFor

\State Assemble
$\widehat{\Theta}=\bigcup_b\widehat{\Theta}_b$
\Comment{Eq.~\ref{eq:overview}}

\Statex
\Statex \textcolor{decodepurple}{\textbf{Stage 2: Deterministic Gaussian rasterization}}

\For{each selected leaf $b$}
    \State Rasterize $\widehat{\Theta}_b$ using the local Gaussian mixture
    \Comment{Eq.~\ref{eq:rasterization}}
\EndFor

\State Assemble the reconstructed blocks and obtain
$\widehat I=\mathcal R(\widehat{\Theta})$
\Comment{Eq.~\ref{eq:overview}}

\State \Return $\widehat I$

\end{algorithmic}
\end{algorithm}

\subsection{More Results}

\begin{figure*}[t]
    \centering
    \includegraphics[width=\linewidth]{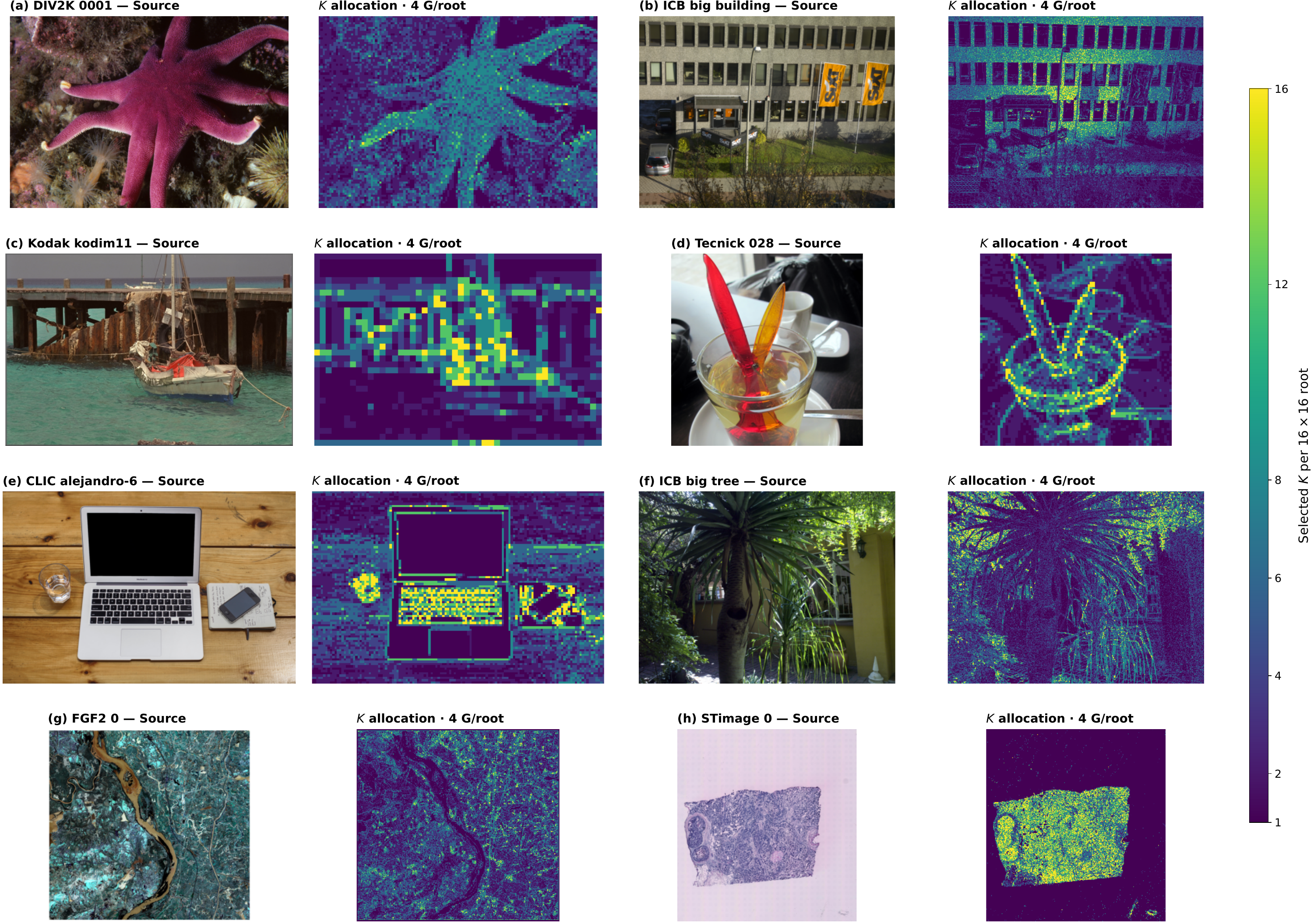}
    \vspace{-20pt}
    \caption{
    Qualitative visualization of Gaussian-budget allocation across eight
    images.
    Each example shows the source image and the selected Gaussian count
    $K_b$ for each $16\times16$ root block under the same average budget
    $B/N=4$.
    Larger $K_b$ values are concentrated in regions with richer local
    structure, while smoother regions receive fewer Gaussians.
    }
    \label{fig:allocation-gallery}
\end{figure*}

\subsubsection{Qualitative allocation analysis.}

\begin{figure*}[t]
    \centering
    \includegraphics[width=\linewidth]{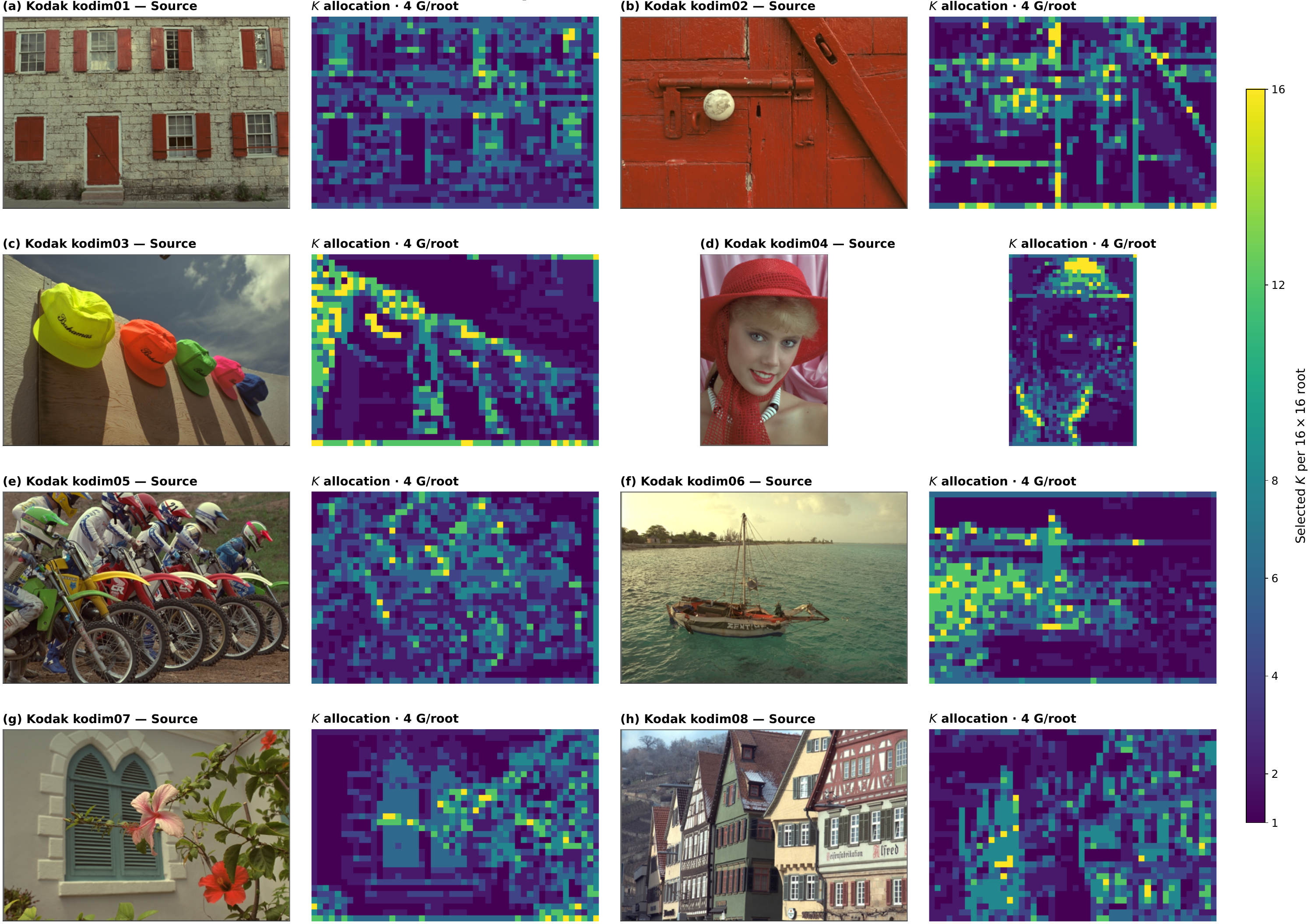}
    \vspace{-20pt}
    \caption{
    More qualitative visualization of Gaussian-budget allocation across eight
    images.
    }
    \label{fig:allocation_gallery_01_kodak}
\end{figure*}

\begin{figure*}[t]
    \centering
    \includegraphics[width=\linewidth]{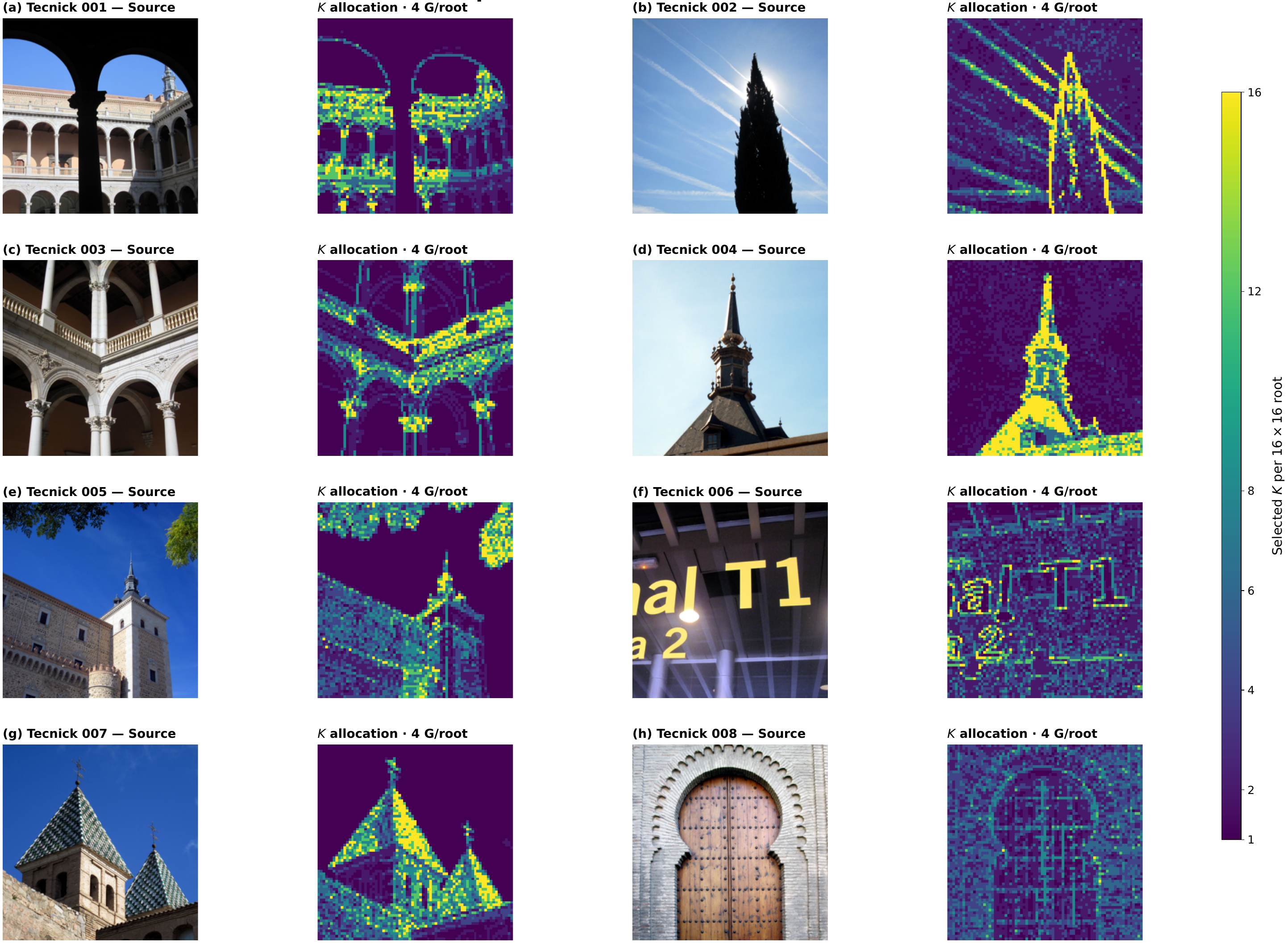}
    \vspace{-20pt}
    \caption{
    More qualitative visualization of Gaussian-budget allocation across eight
    images.
    }
    \label{fig:allocation_gallery_02_tecnick}
\end{figure*}

\begin{figure*}[t]
    \centering
    \includegraphics[width=\linewidth]{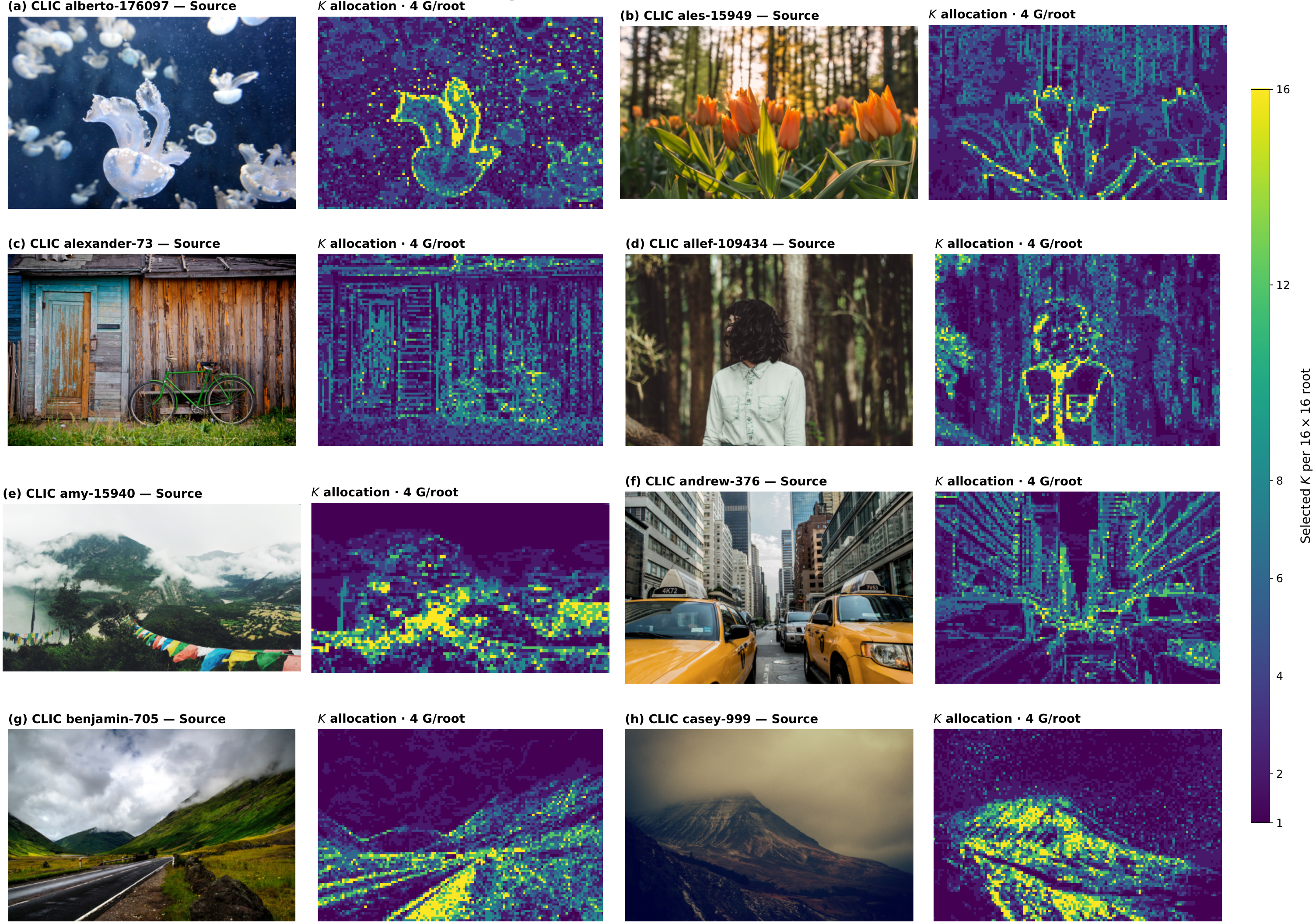}
    \vspace{-20pt}
    \caption{
    More qualitative visualization of Gaussian-budget allocation across eight
    images.
    }
    \label{fig:allocation_gallery_03_clic2020}
\end{figure*}

\begin{figure*}[t]
    \centering
    \includegraphics[width=\linewidth]{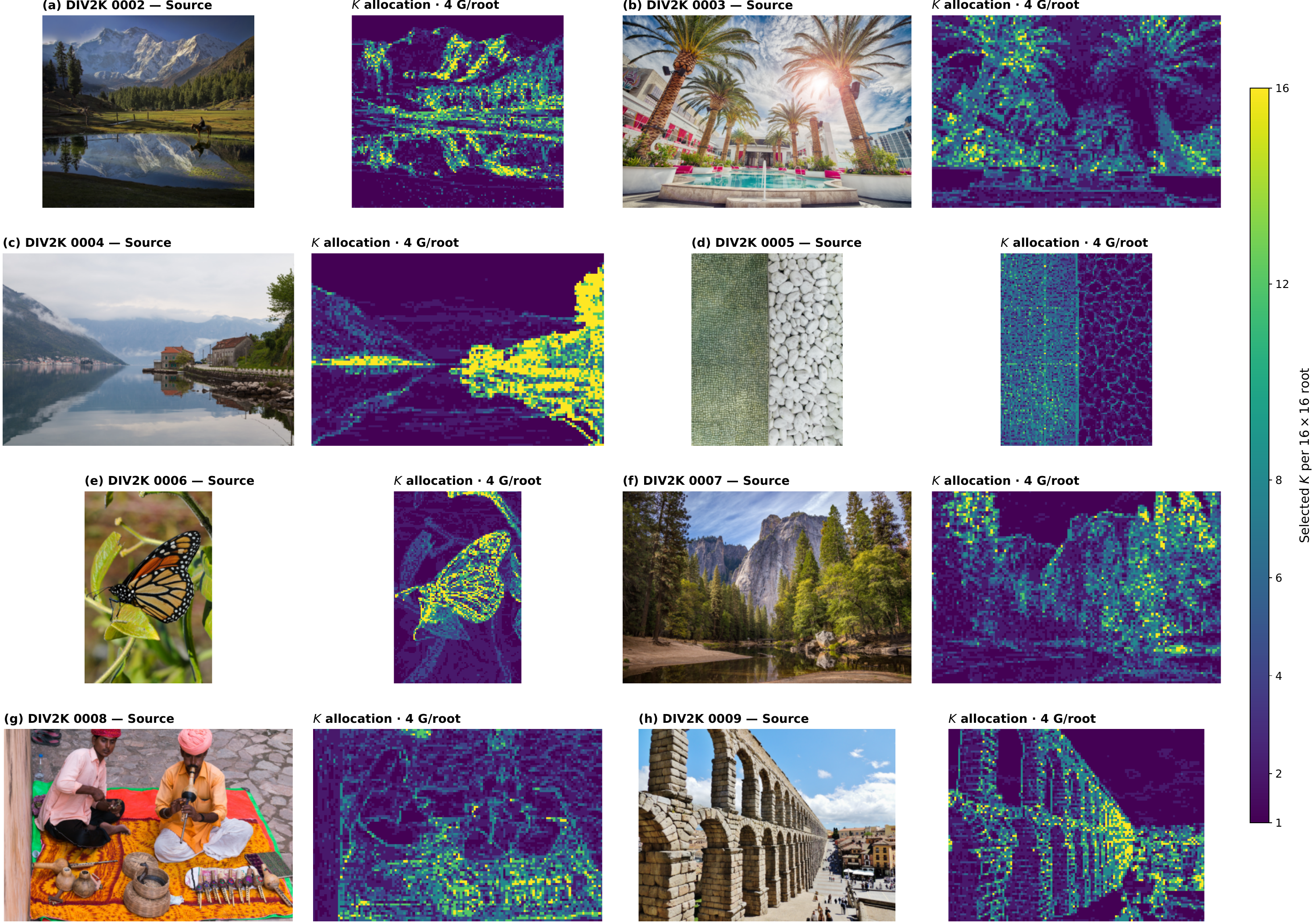}
    \vspace{-20pt}
    \caption{
    More qualitative visualization of Gaussian-budget allocation across eight
    images.
    }
    \label{fig:allocation_gallery_04_div2k}
\end{figure*}

Fig.~\ref{fig:allocation-gallery} examines where the
available Gaussians are placed across different types of images.
Each example contains the source image and its allocation map.
As shonw in this visualization, blocks around locally detailed regions are often assigned larger $K_b$, whereas
smooth backgrounds and nearly empty regions tend to use smaller $K_b$.
For example, detailed building facades and
tissue regions receive more Gaussians than nearby smooth or low-detail
areas.
These results show that the allocator does not distribute the Gaussian
budget uniformly.
Instead, it gives more capacity to regions where additional Gaussians are
more useful for reconstruction.
The visualization supports that our method adapts
both the amount and the spatial placement of Gaussian capacity to the image
content. More results are in Fig.~\ref{fig:allocation_gallery_01_kodak}-\ref{fig:allocation_gallery_04_div2k}.

\begin{table}[t]
\centering
\resizebox{\textwidth}{!}{%
\begin{tabular}{@{}lrrrrrrrr@{}}
\toprule

& \multicolumn{4}{c}{\cellcolor{standardblue}\textbf{Low-rate}}
& \multicolumn{4}{c}{\cellcolor{largeorange}\textbf{High-rate}} \\

\cmidrule(lr){2-5}
\cmidrule(lr){6-9}

\multicolumn{1}{c}{\textbf{Method}} &
\multicolumn{1}{c}{\cellcolor{standardblue}\textbf{\begin{tabular}[c]{@{}c@{}}PSNR\\(dB) $\uparrow$\end{tabular}}} &
\multicolumn{1}{c}{\cellcolor{standardblue}\textbf{MS-SSIM $\uparrow$}} &
\multicolumn{1}{c}{\cellcolor{standardblue}\textbf{\begin{tabular}[c]{@{}c@{}}Training time\\(s) $\downarrow$\end{tabular}}} &
\multicolumn{1}{c}{\cellcolor{standardblue}\textbf{\begin{tabular}[c]{@{}c@{}}Decode speed\\(MP/s) $\uparrow$\end{tabular}}} &
\multicolumn{1}{c}{\cellcolor{largeorange}\textbf{\begin{tabular}[c]{@{}c@{}}PSNR\\(dB) $\uparrow$\end{tabular}}} &
\multicolumn{1}{c}{\cellcolor{largeorange}\textbf{MS-SSIM $\uparrow$}} &
\multicolumn{1}{c}{\cellcolor{largeorange}\textbf{\begin{tabular}[c]{@{}c@{}}Training time\\(s) $\downarrow$\end{tabular}}} &
\multicolumn{1}{c}{\cellcolor{largeorange}\textbf{\begin{tabular}[c]{@{}c@{}}Decode speed\\(MP/s) $\uparrow$\end{tabular}}} \\

\midrule

GLIC
& \textbf{33.34}
& \textbf{0.9655}
& N/A
& 2.67
& \textbf{38.45}
& 0.9874
& N/A
& 2.65 \\

CMIC
& \underline{33.30}
& \underline{0.9654}
& N/A
& \underline{3.44}
& \underline{38.43}
& \underline{0.9890}
& N/A
& \underline{3.43} \\

DCAE
& 33.26
& 0.9593
& N/A
& 3.03
& 38.23
& \textbf{0.9891}
& N/A
& 3.00 \\

\textbf{CaliSplat}
& 27.71
& 0.8866
& \textbf{22.03}
& \textbf{215.82}
& 32.95
& 0.9655
& \textbf{23.84}
& \textbf{175.90} \\

\bottomrule
\end{tabular}%
}

\caption{
Quantitative comparison with representative learned image codecs GLIC~\citep{Chen_2026_CVPR},
CMIC~\citep{chen2026content},
and DCAE~\citep{lu2025learned}.
The best and second-best results for each metric are shown in bold and underlined, respectively.
}
\label{tab:learned_codec_comparison}
\end{table}

\subsubsection{Comparison with learned image codecs.}
Tab.~\ref{tab:learned_codec_comparison} further compares CaliSplat with recent learned image codecs, including
GLIC~\citep{Chen_2026_CVPR},
CMIC~\citep{chen2026content},
and DCAE~\citep{lu2025learned}.
These methods achieve clearly stronger rate--distortion performance, benefiting from dataset-level priors learned offline from large image corpora.
In contrast, Gaussian-based codecs optimize an explicit representation for each individual image and do not directly exploit the same cross-image priors, making this RD gap a broader limitation of the current per-image Gaussian compression paradigm.
CaliSplat nevertheless offers a substantially different decoding trade-off: it reaches 175.90--215.82~MP/s, compared with 2.65--3.44~MP/s for the learned codecs, corresponding to approximately $51.3\times$--$80.8\times$ higher throughput.
Together with Tab.~\ref{tab:quantitative_comparison}, these results show that CaliSplat primarily improves the efficiency of explicit Gaussian compression by reducing per-image fitting cost while preserving its lightweight and highly efficient decoding.

\subsection{Additional Method Details}
\label{app:method_details}

\subsubsection{Gaussian Parameterization and Initialization}
\label{app:param}

For a block of side length $r_b$, pixel indices
$i,j=0,\ldots,r_b-1$ are mapped to normalized local coordinates
$\mathbf p_{ij}=(-1+2i/(r_b-1),\, -1+2j/(r_b-1))$.
For a Gaussian centered at $\boldsymbol\mu=(\mu_y,\mu_x)$, the coordinates along its rotated principal axes are
$d_1=(p_y-\mu_y)\cos\phi+(p_x-\mu_x)\sin\phi$
and
$d_2=-(p_y-\mu_y)\sin\phi+(p_x-\mu_x)\cos\phi$.
These normalized coordinates are used in Eq.~\ref{eq:logit} for all supported block sizes.

During gradient-based per-image fitting, Gaussian parameters are optimized through unconstrained raw variables $\boldsymbol\rho$ and mapped to valid physical values by
\begin{equation}
    \boldsymbol\mu
    =
    1.15\tanh(\boldsymbol\rho_\mu),
    \qquad
    \boldsymbol s
    =
    \operatorname{softplus}(\boldsymbol\rho_s)+0.025,
    \qquad
    \phi=\rho_\phi,\quad
    a=\rho_a,\quad
    \boldsymbol c
    =
    255\,\sigma(\boldsymbol\rho_c).
\end{equation}
The extended center range allows slight motion beyond the normalized block boundary, while the scale floor avoids degenerate ellipses.
Gradients are backpropagated through these mappings and the differentiable Gaussian rasterizer.

For each $(r_b,K)$, we initialize the $K$ Gaussians on the most balanced regular grid satisfying $n_y n_x=K$.
Among valid factor pairs, we choose the one with aspect ratio closest to one; ties and Gaussian ordering follow fixed deterministic rules shared by the encoder and decoder.
For grid location $(i,j)$, the template is
\begin{equation}
    \boldsymbol\mu^{(0)}_{i,j}
    =
    \left(
        \frac{i+\frac12}{n_y^\star}\cdot2-1,\,
        \frac{j+\frac12}{n_x^\star}\cdot2-1
    \right),
    \qquad
    \boldsymbol s^{(0)}
    =
    \left(
        \frac{1.35}{n_y^\star},\,
        \frac{1.35}{n_x^\star}
    \right),
    \qquad
    \phi^{(0)}=0.
\end{equation}
The factor $1.35$ provides overlap between neighboring template Gaussians.
Because the template is determined entirely by $(r_b,K)$, it requires no additional side information.

With the template geometry fixed and initial log-weights set to zero, the rasterizer defines a responsibility matrix
$W\in\mathbb R^{r_b^2\times K}$.
For a flattened RGB block
$X\in\mathbb R^{r_b^2\times3}$,
the initial Gaussian colors are obtained in closed form:
\begin{equation}
    C^{(0)}
    =
    \arg\min_C
    \|WC-X\|_F^2
    =
    W^\dagger X.
\end{equation}
Since $W$ depends only on $(r_b,K)$, its pseudoinverse is reused across blocks with the same configuration.
For $K=1$, the analytic mean is clipped to $[0.5,254.5]$ only when converted to raw sigmoid coordinates, avoiding infinite inverse-sigmoid values; the exact $D_b(1)$ in Eq.~\ref{eq:k1} is computed from the unclipped mean.

\subsubsection{Sparse Candidate Construction}
\label{app:sparse}

For $16\times16$ blocks, we use
\begin{equation}
    \mathcal K_{16}
    =
    \{1,2,4,6,8,12,16\},
    \qquad
    (\rho_1,\rho_2,\rho_4,\rho_6,\rho_8,\rho_{12},\rho_{16})
    =
    (1.0,1.0,0.9,0.8,0.7,0.6,0.5),
\end{equation}
where $\rho_K$ denotes the fraction of blocks retained for candidate size $K$.
For a scale containing $N$ blocks, the corresponding admission count is
\begin{equation}
    M_K
    =
    \begin{cases}
        0, & \rho_K=0,\\[2pt]
        \min\!\left\{
            N,\,
            \max\!\left(
                1,\,
                \left\lceil \rho_K N \right\rceil
            \right)
        \right\},
        & \rho_K>0.
    \end{cases}
\end{equation}
Blocks are ranked by the analytic difficulty $D_b(1)$ as described in Eq.~\ref{eq:gate}; candidates outside $\mathcal A_K$ are treated as unavailable by assigning
$D_b^{\mathrm{cheap}}(K)=+\infty$.

\subsubsection{Teacher Sampling and Calibration}
\label{app:calibration}

Let $N_K=|\mathcal A_K|$ and let
$\sigma_K(0),\ldots,\sigma_K(N_K-1)$ denote the admitted blocks sorted by increasing
$D_b^{\mathrm{cheap}}(K)$.
For each $K>1$, we select at most $128$ teachers, uniformly along this sorted error axis.
With $m_K=\min(128,N_K)$, the sampled set is
\begin{equation}
    \mathcal S_K
    =
    \left\{
        \sigma_K(q_i)
    \right\}_{i=0}^{m_K-1},
    \qquad
    q_i
    =
    \begin{cases}
        0, & m_K=1,\\[2pt]
        \operatorname{round}
        \left(
            \dfrac{i(N_K-1)}{m_K-1}
        \right),
        & m_K>1.
    \end{cases}
    \label{eq:teacher_sampling}
\end{equation}
This sampling covers the cheap-distortion range rather than concentrating teachers around its most populated region.

Each sampled candidate is reinitialized from the same deterministic template used by the cheap fit and optimized independently for $T_t=160$ Adam steps.
We use the same RGB squared-error objective, learning rate, and restart setting as in sparse fitting.
Importantly, teacher optimization does not continue from
$\Theta_{b,K}^{\mathrm{cheap}}$; its purpose is to provide a longer-horizon reference
$D_b^{\mathrm{teach}}(K)$.

For each sampled block, the regression target is
$y_{b,K}=\log(1+D_b^{\mathrm{teach}}(K))$.
Stacking the features $\mathbf x_{b,K}$ from Eq.~\ref{eq:cal_feat} and their targets gives
$X_K$ and $\mathbf y_K$.
We fit a separate least-squares model for each $K$:
\begin{equation}
    \boldsymbol\beta_K
    =
    \arg\min_{\boldsymbol\beta}
    \left\|
        X_K\boldsymbol\beta-\mathbf y_K
    \right\|_2^2.
    \label{eq:calibration_fit}
\end{equation}
The least-squares system is solved directly, without explicitly forming
$(X_K^\top X_K)^{-1}$.

For every admitted candidate, the predicted log-distortion is mapped back to the original distortion domain:
\begin{equation}
    \widetilde D_b(K)
    =
    \max
    \left\{
        \operatorname{expm1}
        \left(
            \mathbf x_{b,K}^{\top}
            \boldsymbol\beta_K
        \right),
        0
    \right\}.
    \label{eq:calibrated_distortion}
\end{equation}
For $K=1$, we retain the exact analytic value
$\widetilde D_b(1)=D_b(1)$, while candidates excluded by the sparse gate remain unavailable with infinite distortion.

\subsubsection{Budget-Aware Hierarchical Allocation}
\label{app:allocation}

For a fixed multiplier $\lambda$, each tree node chooses between remaining unsplit and being replaced by its children.
Let $\mathcal K_n$ denote the legal Gaussian counts at node $n$.
The bottom-up dynamic program is
\begin{equation}
    \begin{aligned}
    F_n^{\mathrm{leaf}}(\lambda)
    &=
    \min_{K\in\mathcal K_n}
    \left[
        \widetilde D_n(K)+\lambda K
    \right],\\
    F_n^{\mathrm{split}}(\lambda)
    &=
    \sum_{c\in\operatorname{child}(n)}
    F_c(\lambda),\\
    F_n(\lambda)
    &=
    \min
    \left\{
        F_n^{\mathrm{leaf}}(\lambda),
        F_n^{\mathrm{split}}(\lambda)
    \right\}.
    \end{aligned}
    \label{eq:tree_dp}
\end{equation}
At an unsplit node, candidate counts are considered in increasing order so that local ties favor the smaller $K$; ties between split and unsplit configurations favor the unsplit node.

For a fixed candidate table and $\lambda$, this recursion gives the optimal legal configuration under the Lagrangian objective.
The argument follows directly by induction on tree depth: terminal nodes choose their best legal candidate, while a split node combines optimal solutions from disjoint child regions and compares their total cost with its own best unsplit candidate.

To meet a target budget $B$, we search over $\lambda$ using the total Gaussian count
$G(\lambda)=\sum_{b\in\mathcal T^\star(\lambda)}K_b^\star(\lambda)$, which is non-increasing in $\lambda$.
Starting from $\lambda_{\mathrm{lo}}=0$ and $\lambda_{\mathrm{hi}}=1$, we double the upper bound until $G(\lambda_{\mathrm{hi}})\le B$, then perform 20 bisection iterations.
Among the feasible configurations near the final boundary, we retain the one with lower calibrated distortion.
Because the candidate space is discrete, varying $\lambda$ need not recover the exact constrained optimum for every integer budget.

\subsubsection{Deterministic Decoding Details}
\label{app:decoder}

The decoder first recovers the selected topology, Gaussian counts, and quantized parameter symbols.
For each leaf, the decoded pair $(r_b,K_b)$ uniquely reconstructs the same deterministic geometric template used by the encoder. Let $q_\mu$ and $q_\Sigma$ denote the quantization factors for center and log-covariance residuals.
The coding-space geometry is recovered as
\begin{equation}
    \widehat{\mathbf g}_{b,k}
    =
    \mathbf g^{(0)}_{b,k}
    +
    \left[
        \frac{z_{\mu_y}}{q_\mu},
        \frac{z_{\mu_x}}{q_\mu},
        \frac{z_m}{q_\Sigma},
        \frac{z_u}{q_\Sigma},
        \frac{z_v}{q_\Sigma}
    \right]^\top .
\end{equation}
Writing
$\alpha=\sqrt{\widehat u^2+\widehat v^2}$,
the anisotropic scale and orientation are recovered by
\begin{equation}
    \widehat\phi
    =
    \begin{cases}
        \frac12\operatorname{atan2}(\widehat v,\widehat u),
        & \alpha>10^{-10},\\
        0, & \text{otherwise},
    \end{cases}
    \qquad
    \widehat s_1
    =
    2^{\widehat m+\alpha},
    \qquad
    \widehat s_2
    =
    2^{\widehat m-\alpha}.
\end{equation}
Relative weights and palette-coded colors are inverted analogously, yielding the explicit Gaussian representation $\widehat{\Theta}$. Each selected leaf is then rasterized using the same Gaussian formulation as in Eqs.~\ref{eq:logit}--\ref{eq:rasterization}.
The reconstructed image is rounded and clipped to RGB8, assembled over all leaves, and cropped to the original $H\times W$ resolution.

\subsection{Implementation Details}
\label{app:implementation}

The global Gaussian budget is determined using 20 bisection iterations over
the Lagrangian multiplier.
For parameter coding, we use quantization denominators of 20, 10, and 3 for
Gaussian centers, log-covariance parameters, and relative weights,
respectively, together with a color quantization step of 2.
The color palette contains at most 768 entries and is optimized with up to four
Lloyd iterations.

\newpage

\begin{figure}[t]
    \centering
    \includegraphics[width=\linewidth]{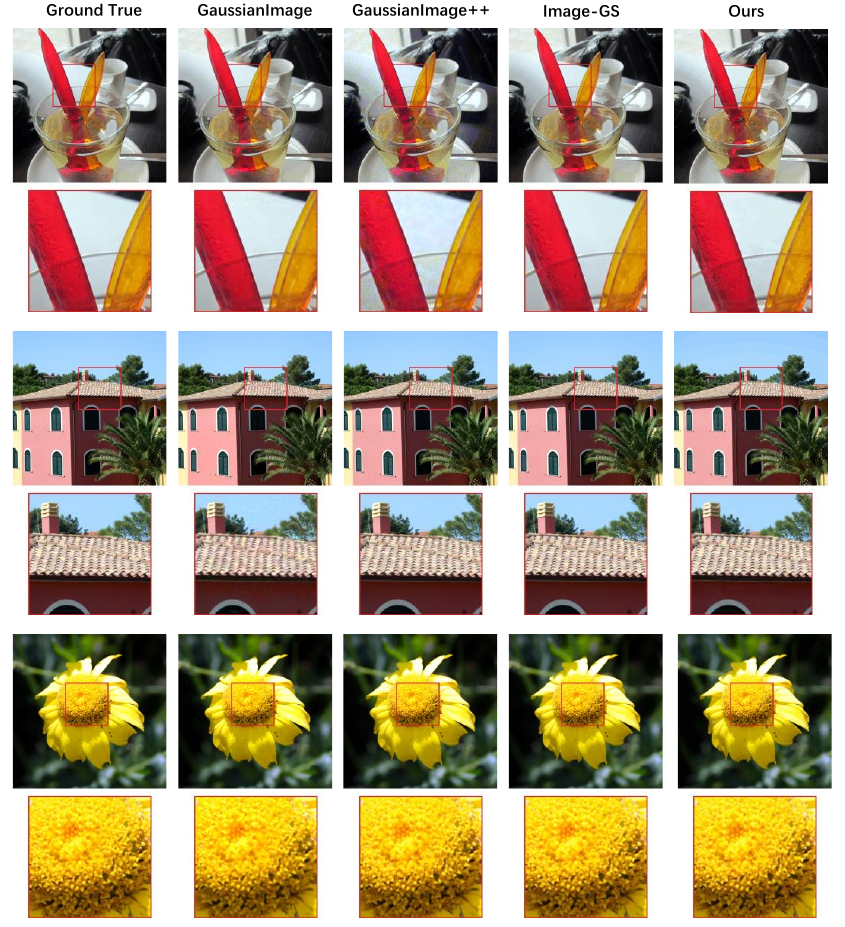}
\caption{
More qualitative reconstruction results of CaliSplat with other baselines.
}
    \label{fig:vis1}
\end{figure}

\begin{figure}[t]
    \centering
    \includegraphics[width=\linewidth]{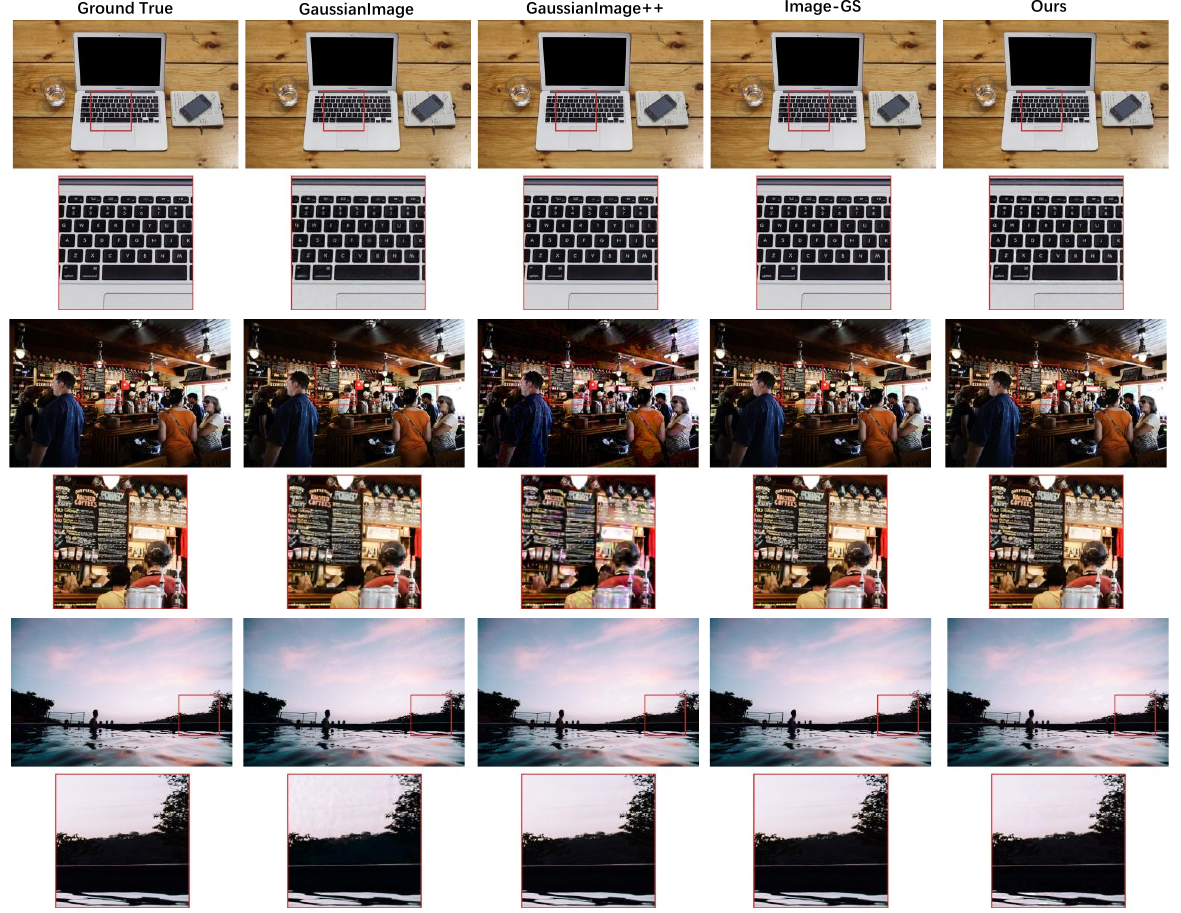}
\caption{
More qualitative reconstruction results of CaliSplat with other baselines.
}
    \label{fig:vis2}
\end{figure}

\begin{figure}[t]
    \centering
    \includegraphics[width=\linewidth]{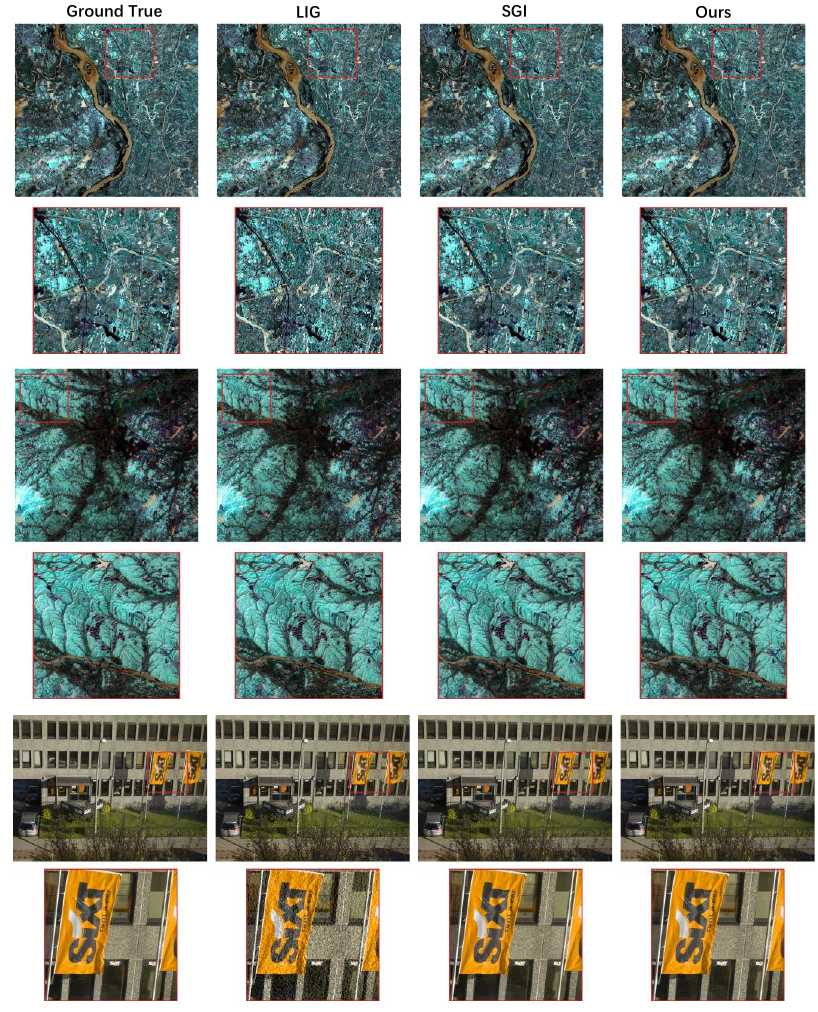}
\caption{
More qualitative reconstruction results of CaliSplat with other baselines.
}
    \label{fig:vis3}
\end{figure}

\begin{figure}[t]
    \centering
    \includegraphics[width=\linewidth]{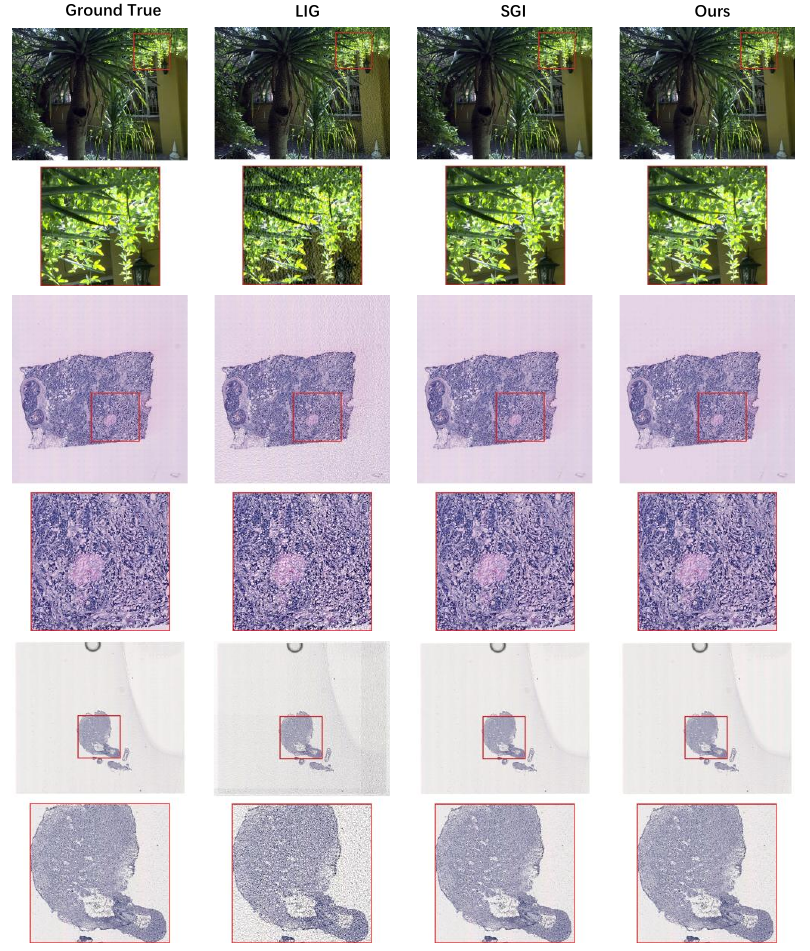}
\caption{
More qualitative reconstruction results of CaliSplat with other baselines.
}
    \label{fig:vis4}
\end{figure}

\begin{table*}[t]
\centering
\caption{Main notations used in CaliSplat.}
\label{tab:notations}
\renewcommand{\arraystretch}{1.08}
\setlength{\tabcolsep}{6pt}
\begin{tabularx}{\textwidth}{@{}lX@{}}
\toprule
\textbf{Notation} & \textbf{Description} \\
\midrule

$I$, $\widehat I$
& Input and reconstructed RGB images. \\

$S$
& Self-contained compressed bitstream. \\

$b$
& Spatial image block. \\

$r_b$
& Side length of block $b$. \\

$K$
& Number of Gaussians used by a candidate representation. \\

$k$
& Index of a Gaussian within a candidate. \\

$\theta_{b,k}$
& Parameters of the $k$-th Gaussian in block $b$. \\

$\Theta_{b,K}$
& $K$-Gaussian candidate representation for block $b$. \\

$\mathcal R(\cdot)$
& Deterministic local Gaussian rasterization operator. \\

$D_b(1)$
& Analytic $K=1$ reconstruction error used as a block-difficulty signal. \\

$\mathcal A_K$
& Set of blocks admitted for fitting a $K$-Gaussian candidate. \\

$C^{(0)}$
& Closed-form color initialization for a Gaussian candidate. \\

$T_c$
& Number of optimization steps used for cheap candidate fitting. \\

$\Theta_{b,K}^{\mathrm{cheap}}$
& Gaussian parameters obtained from short-horizon candidate fitting. \\

$D_b^{\mathrm{cheap}}(K)$
& Reconstruction error of the cheap-fitted candidate. \\

$D_b^\star(K)$
& Reconstruction error associated with sufficiently long optimization. \\

$\varepsilon_{b,K}$
& Residual optimization error caused by early stopping. \\

$T_t$
& Number of optimization steps used for sampled teacher fitting. \\

$D_b^{\mathrm{teach}}(K)$
& Longer-horizon reconstruction error produced by a sampled teacher fit. \\

$\mathbf x_{b,K}$
& Feature vector used for per-$K$ score calibration. \\

$\widetilde D_b(K)$
& Calibrated candidate score used for global Gaussian allocation. \\

$B$
& Global Gaussian budget. \\

$\mathcal T$
& Selected block structure under the global budget. \\

$K_b$
& Number of Gaussians assigned to selected block $b$. \\

$\lambda$
& Cost assigned to each Gaussian during budget allocation. \\

$\mathbf z_{b,k}$
& Coding-space geometry representation of a selected Gaussian. \\

$\mathbf z^{(0)}_{b,k}$
& Decoder-known geometry of the deterministic template. \\

$\Delta\mathbf z_{b,k}$
& Template-relative geometry residual,
$\mathbf z_{b,k}-\mathbf z^{(0)}_{b,k}$. \\

$\widehat{\Theta}$
& Explicit Gaussian representation recovered by the decoder. \\

\bottomrule
\end{tabularx}
\end{table*}

\end{document}

%% file: math_commands.tex
\usepackage{amsmath,amsfonts,bm}

\def\eqref#1{equation~\ref{#1}}

\def\1{\bm{1}}

\DeclareMathAlphabet{\mathsfit}{\encodingdefault}{\sfdefault}{m}{sl}
\SetMathAlphabet{\mathsfit}{bold}{\encodingdefault}{\sfdefault}{bx}{n}

